\documentclass{IEEEtran}%

\usepackage[numbers]{natbib}
\usepackage{algorithm}%
\usepackage{algorithmic}%
\usepackage{amssymb,amsmath}%
\usepackage{url}%
\usepackage{subfigure}%
\usepackage{graphicx}%
\usepackage{setspace}

\title{Discrete and Continuous, Probabilistic Anticipation for Autonomous Robots in Urban Environments}%
\author{Frank Havlak and Mark Campbell
\thanks{F. Havlak and M. Campbell are with the Sibley School of Mechanical and Aerospace Engineering, Cornell University, Ithaca, NY 14853}%
\thanks{For further information, send correspondence to author Frank Havlak. E-mail: fh92@cornell.edu}}%

\begin{document}

\maketitle%

\begin{abstract}

This paper develops a probabilistic anticipation algorithm for dynamic objects observed by an autonomous robot in an urban environment. Predictive Gaussian mixture models are used due to their ability to probabilistically capture continuous and discrete obstacle decisions and behaviors; the predictive system uses the probabilistic output (state estimate and covariance) of a tracking system, and map of the environment to compute the probability distribution over future obstacle states for a specified anticipation horizon. A Gaussian splitting method is proposed based on the sigma-point transform and the nonlinear dynamics function, which enables increased accuracy as the number of mixands grows. An approach to caching elements of this optimal splitting method is proposed, in order to enable real-time implementation. Simulation results and evaluations on data from the research community demonstrate that the proposed algorithm can accurately anticipate the probability distributions over future states of 
nonlinear systems.

\end{abstract}

\section{Introduction}
Autonomous urban driving is an important and maturing field in mobile robotics. Intelligent vehicles promise to improve both road safety, vehicle efficiency, and convenience\cite{folsom2011social,bishop2000intelligent,juan2006socio}. The finals of the 2007 DARPA Urban Challenge (DUC) was an empirical evaluation of the state-of-the-art at the time, integrating 11 autonomous vehicles together with other robots and human drivers in an urban environment for long-duration operations ($>$ 4 hours)\cite{miller2008team,urmson2008autonomous,bacha2008odin}. Continued development in the field has led to autonomous cars beginning to drive in real urban environments alongside civilian traffic \cite{markoff2012collision, levinson2011towards, fuberlin_pr, markoff2010google}.

Many autonomous cars use primarily reactionary planners that rely on rapid re-planning in order to respond to the dynamic environments in which they operate \cite{huttenlocher2008team}. A collision between the MIT and Cornell entries was one of several examples in the 2007 DUC that raised safety concerns about reactionary planning for autonomous driving \cite{fletcher2008cornell}. One approach proposed to more intelligently handle autonomous driving in dynamic environments is to incorporate `anticipation' into path planning, or predicting the future motion of dynamic obstacles for use in planning. This area has been an active topic in mobile robotics in recent years \cite{hardy2010iros,ziebart2009planning}. This paper presents a formal method for probabilistically anticipating the future behavior of tracked obstacles.

The problem of anticipation is inherently probabilistic, as it is impossible to know the true future behavior of dynamic obstacles that make their own independent decisions. In addition, the behavioral models used to anticipate obstacle behavior are often highly non-linear. In the literature, several algorithms have been proposed to simplify the problem, such as assuming no uncertainty in future obstacle behavior \cite{Petti05safemotion,choi2010analytic,ohki2010collision}. These algorithms are well suited for cooperative situations, where the obstacles can communicate their intentions to the robot, or for short anticipation horizons. However, they do not provide sufficient information for reasoning about an obstacle with unknown intentions over a significant anticipation horizon. Similarly, several proposed methods consider only a subset of obstacle uncertainty, such as along-track error \cite{miura2002probabilistic}. These approaches reduce the complexity of the problem to a manageable level, while 
still considering the probabilistic aspects of obstacle anticipation, but are typically very simple and narrow in their application.

Another class of algorithms applies standard estimation filters (Kalman Filter, Extended Kalman Filter, etc.) to the problem of anticipation \cite{fulgenzi2009probabilistic, du-robotic}. Such approaches assume a model for the behavior of the obstacle, and provide mathematically rigorous, probabilistic estimates of that obstacle's state over the anticipation horizon. These approaches are well-suited to obstacles that are accurately described by linear models because they maintain a single Gaussian to represent the uncertainty. For obstacles with more complex behaviors, such as those based on non-linear dynamics (e.g. cars, bicycles, etc.) and those that make discrete decisions (e.g. intersections, passing, etc.), the uncertainty of the anticipated obstacle state becomes inaccurate very quickly, thus severely limiting the prediction horizon. \citet{du-robotic} has been extended by the authors to use Gaussian Mixture Models (GMMs) to capture multiple obstacle hypothesis, but is still focused primarily on linear 
obstacle models \cite{dutoit2012robot}.

More complex uncertainties can be addressed, while avoiding the linearization problems of standard filters; for example, Monte-Carlo (MC) methods\cite{ferguson2008detection}. These approaches are attractive because they can consider complex, non-Gaussian uncertainties and allow for the use of non-linear obstacle models to capture complex obstacle behavior. However, the accuracy of prediction scales with the number of particles, and there are no guarantees that the particle set effectively covers the true distribution of possible future obstacle states. Because the assumed dynamics model for the obstacle has to be evaluated at every particle used in anticipation, increasing confidence in the estimate is strongly traded with computational resources.

A GMM based predictor is proposed in the paper to anticipate obstacle behaviors \cite{alspach1972nonlinear}. GMMs provide a well-suited representation to probabilistically anticipate non-Gaussian obstacle states. Here, the GMM is used to uniquely include discrete state elements that capture complex, high-level obstacle behaviors. Accurate anticipation of a wide variety of dynamic obstacles is ensured using a novel method for detecting linearization errors using sigma-point methods, and adjusting the mixture accordingly by optimally splitting inaccurately propagated mixands. This approach reduces the individual covariances of inaccurately propagated mixands, bringing them into a nearly linear regime. The presented algorithm provides accurate, probabilistic future obstacle state estimates and is shown to perform well even with highly non-linear obstacle motion models.

This paper is outlined as follows: Section~\ref{sec:obs_st} defines the representation of the obstacle state and Section~\ref{sec:hmm} defines the mixture model. Sections~\ref{sec:prop} describes the discrete and continuous anticipation of the obstacle state. Section~\ref{sec:overview} provides an overview of the proposed algorithm. The details of temporal propagation are described in Section~\ref{sec:cont_prop}, including the non-linearity detection and mixand splitting. Section~\ref{sec:example_sim} provides an example implementation of the anticipation algorithm in simulation, and Section~\ref{sec:clif_data} demonstrates the efficacy of the algorithm on a real data set. Section~\ref{sec:MITCollision} demonstrates the potential safety improvements by applying the proposed algorithm to the 2007 MIT-Cornell autonomous collision.

\section{Hybrid Mixture Anticipation Algorithm}
\label{sec:algo}

The hybrid mixture anticipation algorithm described in this paper is designed to predict the probability distribution over the state of a tracked obstacle forward in time. Hybrid, here, is used to denote jointly discrete and continuous components. The intended application of the presented algorithm is to make accurate, probabilistic information about future obstacle behaviors available for use in path planning. In order to provide the most general algorithm, obstacle models can include non-linear behaviors, as well as discrete variables to capture higher-level decisions. To meet these requirements, the distribution over the obstacle state is described using a hybrid Gaussian/discrete mixture model (hGMM).

\subsection{Obstacle State Model}
\label{sec:obs_st}
The obstacle state at time $k$ ($\mathbf{x}_k$) is assumed to have continuous elements ($\mathbf{x}_k^{\text{C}}$, representing position, velocity, etc.) and discrete elements ($\mathbf{x}_k^{\text{D}}$, representing behavioral modes). The state vector is partitioned accordingly:

\begin{align}
 \mathbf{x}_k &= \begin{bmatrix}
                  \mathbf{x}_k^{\text{C}}\\
		  \mathbf{x}_k^{\text{D}}
                 \end{bmatrix}.
\end{align}
\noindent
A model for the evolution of the obstacle state is assumed to be available, and partitioned into discrete ($f^{\text{D}}$) and continuous ($f^{\text{C}}$) components:

\begin{align}
 \label{eq:dyn_mdl}
 \mathbf{x}_{k+1}^{\text{D}} &= f^{\text{D}}\left(\mathbf{x}_{k}^{\text{D}},\mathbf{x}_{k}^{\text{C}},\mathbf{v}_k\right) \nonumber \\
 \mathbf{x}_{k+1}^{\text{C}} &= f^{\text{C}}\left(\mathbf{x}_{k+1}^{\text{D}},\mathbf{x}_{k}^{\text{C}},\mathbf{v}_k\right)
\end{align}

\noindent
where $\mathbf{v}_k$ is the process noise vector at time $k$. $f^{\text{D}}$ and $f^{\text{C}}$ are both functions that take as inputs \emph{point values} for the vehicle state ($\mathbf{x}_{k}^{\text{D}},\mathbf{x}_{k}^{\text{C}}$) and process noise ($\mathbf{v}_k$), as this is typically how models are defined. The following sections generalize these to $h^{\text{D}}$ and $h^{\text{C}}$, which operate over distributions over, rather than samples from, the vehicle state and process noise.

\subsection{Hybrid Mixture Probability Distribution Representation}
\label{sec:hmm}
The probability distribution $p(\mathbf{x}_k)$ is approximated using an hGMM. The hGMM extends the GMM presented in \citet{alspach1972nonlinear} by including discrete variables. Discrete variables allow the hGMM to capture both continuous behavior of a system (position, velocity, etc.) as well as high-level, abstract behaviors (turning left, going straight, etc.). The hGMM inherits the capability of GMMs to represent many general probability distributions, with increasing accuracy in the limit of a large number of mixands, while still maintaining the convenient computational properties of Gaussian distributions. The hGMM is defined:

\begin{align}
 \label{eq:mixture}
 p(\mathbf{x}_k) &= \sum_{i=1}^{M_k}w_k^i\cdot p^i(\mathbf{x}_k) \nonumber
\end{align}

\noindent
where $M_k$ is the number of mixands in the hGMM at time $k$, and $w_k^i$ are the weights on each mixand at time $k$, such that

\begin{align}
 \sum_{i=1}^{M_k}w_k^i &= 1, \text{ and } w_k^i > 0 \quad \forall i.
\end{align}

The mixand $p^i(\mathbf{x}_k)$ is defined as a Gaussian distribution over the continuous state elements and a hypothesis (using a delta function) over the discrete state elements:

\begin{align}
 \label{eq:mixand}
 p^i(\mathbf{x}_k) &= \delta(\mathbf{x}_k^{\text{D}} - \alpha_k^{i})\cdot\mathcal{N}(\mathbf{x}_k^{\text{C}}\vert\mu_k^{i},\Sigma_k^i)
\end{align}

\noindent
where $\alpha_k^{i}$ is the mixand hypothesis of the discrete state, and $\mu_k^{i}$ and $\Sigma_k^i$ are the mean and covariance of the Gaussian distribution over the continuous state.

\subsection{Hybrid Mixture Propagation}
\label{sec:prop}
The hGMM formulation enables each mixand to be propagated forward in time independently using the dynamics model (Equation~\ref{eq:dyn_mdl}), thereby reducing the complexity of the problem. The propagation of the mixands is complicated in two ways, however. First, more than one discrete state can be transitioned into (for example, a tracked obstacle approaching an intersection may turn left, turn right, or continue straight). Second, the variance on the continuous state elements may grow to the point where the mixand can no longer be accurately propagated through the dynamics model. Each of these are addressed in the proposed probabilistic anticipation algorithm.

Consider the discrete mixand propagation through the dynamics function from Equation~\ref{eq:dyn_mdl}, is:

\begin{multline}
 \begin{bmatrix}
  \left(\alpha_k^{1},\mu_k^{1},\Sigma_k^1, w_k^1\right) \\
  \vdots \\
  \left(\alpha_k^{M_k},\mu_k^{M_k},\Sigma_k^{M_k}, w_k^{M_k}\right)
 \end{bmatrix} \overset{h^{\text{D}}}{\rightarrow}\\
                                               \begin{bmatrix}
					        \left(\alpha_{k+1}^{1},\mu_{k-}^{1},\Sigma_{k-}^1,w_{k-}^1\right) \\
				   	        \vdots \\
					        \left(\alpha_{k+1}^{M_{k-}},\mu_{k-}^{M_{k-}},\Sigma_{k-}^{M_{k-}}, w_{k-}^{M_{k-}}\right)
                                               \end{bmatrix}
\end{multline}

\noindent
where $\alpha_{k+1}^i$ is the discrete state for the $i^{th}$ mixand at the next time step ($k+1$). In cases where a mixand has multiple possible next discrete states (such as a choice of roads at an intersection), the mixand is split so that one copy of the mixand can transition to each possible discrete state. Although the continuous aspects of the mixand ($\mu_k^{i},\Sigma_k^i$) are not affected by the discrete propagation, the time index $k-$ is used to account for growth in the mixture size due to mixands transitioning to multiple discrete states.

Similarly, the continuous mixand propagation function $h^{\text{C}}$, is defined as:

\begin{multline}
 \label{eq:c_funcs}
 \begin{bmatrix}
  \left(\alpha_{k+1}^{1},\mu_{k-}^{1},\Sigma_{k-}^1, w_{k-}^1\right) \\
  \vdots \\
  \left(\alpha_{k+1}^{M_{k-}},\mu_{k-}^{M_{k-}},\Sigma_{k-}^{M_{k-}}, w_{k-}^{M_{k-}}\right)
 \end{bmatrix} \overset{h^{\text{C}}}{\rightarrow} \\
                                               \begin{bmatrix}
					        \left(\alpha_{k+1}^{1},\mu_{k+1}^{1},\Sigma_{k+1}^1,w_{k+1}^1\right) \\
				   	        \vdots \\
					        \left(\alpha_{k+1}^{M_{k+1}},\mu_{k+1}^{M_{k+1}},\Sigma_{k+1}^{M_{k+1}}, w_{k+1}^{M_{k+1}}\right)
                                               \end{bmatrix}
\end{multline}

\noindent
where the right-hand side characterizes the propagated hGMM after one full step.

\subsection{Algorithm Overview}
\label{sec:overview}
\begin{figure}[!t]
 \centering
  \includegraphics[scale=.2]{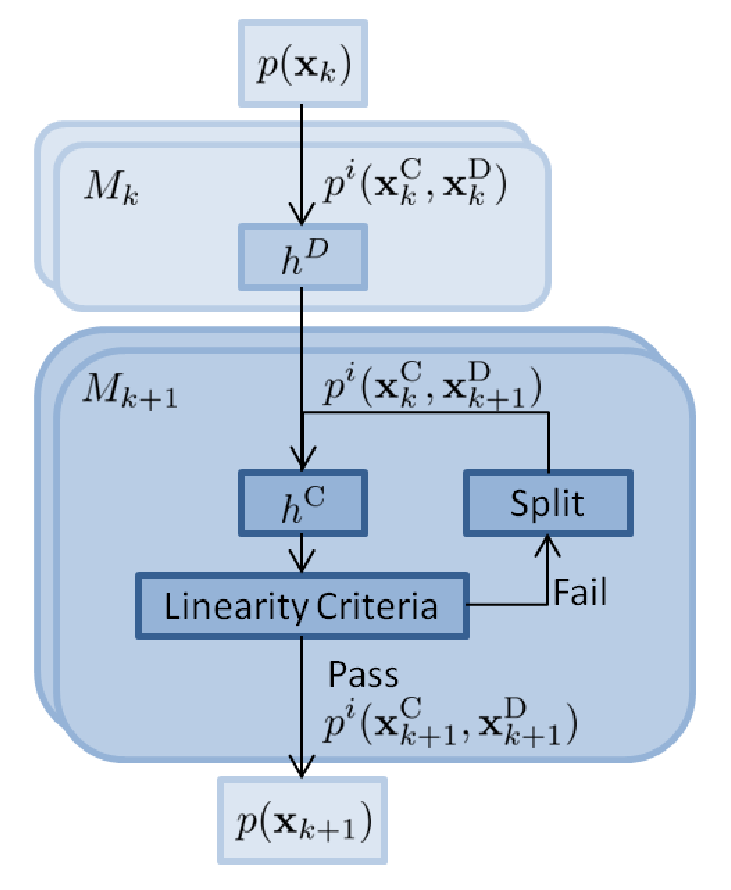}
 \caption{Block diagram illustrating propagation of hybrid Gaussian mixture through a given dynamics model}
 \label{fig:block_diagram}
\end{figure}

The propagation steps of the hGMM are summarized in Figure~\ref{fig:block_diagram}. First, each mixand $p^i(\mathbf{x}_k)$ in the hGMM at time $k$ is propagated through the discrete dynamics. This step has the potential to increase the number of mixands in the hGMM due to the possibility of multiple discrete decisions being available to mixands. Next, each mixand in the hGMM after discrete propagation is propagated through the continuous dynamics. This step includes a linearity criteria (Section~\ref{sec:nonlin}) that ensures the accuracy of propagation by splitting mixands that propagate inaccurately (Section~\ref{sec:split}). After the continuous propagation is completed, the probability distribution over the state at time $k+1$ can be written as:

\begin{multline}
 \label{eq:propmix}
 p(\mathbf{x}_{k+1}) = \sum_{i=1}^{M_{k+1}}w_{k+1}^i\cdot \delta(\mathbf{x}_{k+1}^{\text{D}} - \alpha_{k+1}^{i})\cdot \\
\mathcal{N}(\mathbf{x}_{k+1}^{\text{C}}\vert\mu_{k+1}^{i},\Sigma_{k+1}^i).
\end{multline}

\noindent
Note that the number of mixands ($M_{k+1}$) in the hGMM after the continuous propagation can be larger than the number of mixands after the discrete propagation ($M_{k-}$) due to the possibility that some mixands were split to ensure accurate propagation through nonlinear dynamics.

\section{Mixand Continuous Propagation}
\label{sec:cont_prop}
The first step in the mixture continuous propagation is to predict the continuous state distribution in each of the mixands forward one time step. The continuous dynamics function $h^{\text{C}}$ (Equation~\ref{eq:c_funcs}) uses the sigma-points (SP) transform (sometimes called the unscented transform) \cite{julier1997new, uhlmann1995dynamic, julier1996general, wan2000unscented} to propagate the mixand through the continuous dynamics function $f^{\text{C}}$ (Equation~\ref{eq:dyn_mdl}). Sigma-points are well-explored for use in estimation problems involving nonlinear dynamics and measurement models\cite{julier1997new,julier1996general}. The Gaussian distributions $\mathcal{N}(\mathbf{x}_k^{\text{C}}\vert\mu_k^{i},\Sigma_k^i)$ and $\mathcal{N}(\mathbf{v}_k\vert\boldsymbol{0},\Sigma_{\mathbf{v}})$ are approximated by deterministically choosing sets of points $\boldsymbol{\chi}_{k}^i$ and $\boldsymbol{\Upsilon}_k$, called sigma points: 

\begin{align}
 \label{eq:sigpts}
 \boldsymbol{\chi}_{k}^i =& \begin{bmatrix}
                             \chi_k^{i,1},...,\chi_k^{i,1+2n_{\mathbf{x}}+2n_{\mathbf{v}}}
                            \end{bmatrix} \nonumber \\
 \boldsymbol{\Upsilon} =& \begin{bmatrix}
                           \Upsilon^{1},...,\Upsilon^{1+2n_{\mathbf{x}}+2n_{\mathbf{v}}}
                          \end{bmatrix}
\end{align}

\noindent
where $n_{\mathbf{x}}$ is the dimension of $\mathbf{x}_k^{\text{C}}$ and $n_{\mathbf{v}}$ is the dimension of $\mathbf{v}_k$; the process noise distribution is assumed to be time-invariant and Gaussian without loss of generality -- time varying GMM process noise distributions can be used with minor modifications. To find the individual sigma points in Equation~\ref{eq:sigpts}, the matrix square-roots of the covariances $\Sigma_k^i$ and $\Sigma_{\mathbf{v}}$ are first solved:

\begin{align}
 \mathbf{S}_{\mathbf{x},k}^i\cdot\left(\mathbf{S}_{\mathbf{x},k}^{i}\right)^T &= \Sigma_k^i \nonumber \\
 \mathbf{S}_{\mathbf{v}}\cdot\left(\mathbf{S}_{\mathbf{v}}\right)^T &= \Sigma_{\mathbf{v}} \nonumber
\end{align}

\noindent
such that

\begin{align}
 \mathbf{S}_{\mathbf{x},k}^i &= \begin{bmatrix}
                                 S_{\mathbf{x},k}^{i,1},...,S_{\mathbf{x},k}^{i,n_{\mathbf{x}}}
                                \end{bmatrix} \nonumber \\
 \mathbf{S}_{\mathbf{v}} &= \begin{bmatrix}
                             S_{\mathbf{v}}^{1},...,S_{\mathbf{v}}^{n_{\mathbf{v}}}
                            \end{bmatrix}.
\end{align}
\noindent
The individual sigma-points are then defined using these matrix square-roots and a parameter $\lambda$:

\begin{align}
 \chi_k^{i,j} &= \begin{cases}
                 \mu_k^{i}, & j \in \{0,[2n_{\mathbf{x}} + 1,2n_{\mathbf{x}} + 2n_{\mathbf{v}}]\} \\
		 \mu_k^{i} + \gamma\cdot S^{i,j}_{\mathbf{x},k}, & j \in [j=1,n_{\mathbf{x}}] \\
                 \mu_k^{i} - \gamma\cdot S^{i,j-n_{\mathbf{x}}}_{\mathbf{x},k}, & j \in [n_{\mathbf{x}}+1,2n_{\mathbf{x}}] \\
                \end{cases} \nonumber \\
 \Upsilon^j &= \begin{cases}
                \boldsymbol{0}, & j\in[0,2n_{\mathbf{x}}] \\
	        \gamma\cdot\mathbf{S}_{\mathbf{v},k}^{j-2n_{\mathbf{x}}}, &  j\in[2n_{\mathbf{x}}+1,2n_{\mathbf{x}}+n_{\mathbf{v}}] \\
                -\gamma\cdot\mathbf{S}_{\mathbf{v},k}^{j-2n_{\mathbf{x}}-n_{\mathbf{v}}}, & j\in[2n_{\mathbf{x}}+n_{\mathbf{v}}+1,2n_{\mathbf{x}}+2n_{\mathbf{v}}] \\
               \end{cases}
\end{align}
\noindent
where $\gamma = \sqrt{n_{\mathbf{x}}+n_{\mathbf{v}}+\lambda}$.

Each pair of points ($\chi_k^{i,j}, \Upsilon^j$) is then individually propagated through $\chi_{k+1}^{i,j} = f^{\text{C}}\left(\alpha^{i}_{k+1},\chi_{k}^{i,j},\Upsilon^j\right)$, and the resulting set $\boldsymbol{\chi}_{k+1}^i = \left[\chi_{k+1}^{i,1},...,\chi_{k+1}^{i,1+2n_{\mathbf{x}}+2n_{\mathbf{v}}}\right]$ is used to evaluate the impact of nonlinearities (Section~\ref{sec:nonlin}) on the accuracy of prediction, and to find the predicted distribution at time $k+1$, $\mathcal{N}(\mathbf{x}_{k+1}^{\text{C}}\vert\mu_{k+1}^{i},\Sigma_{k+1}^i)$ (Section~\ref{sec:nosplit}).

\subsection{No Splitting Case}
\label{sec:nosplit}
If the mixand passes the linearity criteria described in Section~\ref{sec:nonlin}, the propagated sigma points are used to find the predicted mean and covariance ($\mu^{i}_{k+1}$ and $\Sigma^i_{k+1}$) for the mixand:

\begin{align}
 \label{eq:sigma_combine}
 \mu^{i}_{k+1} &= \sum_{j=1}^{1+2n_{\mathbf{x}}+2n_{\mathbf{v}}}\gamma^j \cdot \chi_{k+1}^{i,j} \nonumber \\
 \Sigma^{i}_{k+1} &= \sum_{j=1}^{1+2n_{\mathbf{x}}+2n_{\mathbf{v}}}\beta^j \cdot \left( \chi_{k+1}^{i,j} - \mu^{i}_{k+1} \right)\left( \chi_{k+1}^{i,j} - \mu^{i}_{k+1} \right)^{T} \nonumber \\
 \gamma^j &= \begin{cases}
              \frac{\lambda}{\lambda + n_{\mathbf{x}} + n_{\mathbf{v}}} &\text{for $j=1$}\\
              \frac{1}{2\left(\lambda + n_{\mathbf{x}} + n_{\mathbf{v}}\right)} &\text{for $j \neq 1$}\\
             \end{cases} \nonumber \\
 \beta^j &= \begin{cases}
             \frac{\lambda}{\lambda + n_{\mathbf{x}} + n_{\mathbf{v}}}+2 &\text{for $j=1$}\\
             \frac{1}{2\left(\lambda + n_{\mathbf{x}} + n_{\mathbf{v}}\right)} &\text{for $j \neq 1$}\\
            \end{cases}.
\end{align}

Once the mean and covariance are computed, the propagation of the mixand is complete. If the mixand fails the linearity criteria, it is split into several mixands with reduced covariance, described in Section~\ref{sec:split}.

\subsection{Linearity Criteria}
\label{sec:nonlin}
The strength of Gaussian mixture models is that they can accurately approximate non-Gaussian pdfs that arise either through non-Gaussian process or measurement noise or through non-linearities in the system model\cite{alspach1972nonlinear}. Traditional Gaussian mixture model propagation algorithms \cite{simandl2005sigma,sorenson1971recursive} use standard non-linear filtering techniques, such as Extended Kalman Filtering or Unscented Kalman Filtering, to propagate the individual mixands without considering the impact of local non-linearities in the system model on the accuracy of propagation. Motivated by a careful examination of the sigma-points, this paper develops a natural measure for how accurately a given mixand is propagated through the system model, along with a splitting method to reduce covariances, which, in turn, improves propagation accuracy.

The SP transform propagates Gaussian mixands through a temporal model using sigma-points; intuitively, this process performs a statistical linearization of the model in the neighborhood of the Gaussian mixand. However, the SP transform does not provide a measure of error introduced by the statistical linearization. In order to evaluate the error introduced by the SP propagation, the linearization residuals are proposed here as a metric. The process is as follows. First, a linear model is defined:

\begin{align}
 \label{eq:linearization}
 \bar{\boldsymbol{\chi}}_{k+1}^i &= \mathbf{A}\bar{\boldsymbol{\chi}}_{k}^i + \mathbf{b}\cdot\boldsymbol{1}_n \nonumber \\
                          &= \left[\mathbf{A},\mathbf{b}\right] \left[\begin{array}{l l}
                                                                       \bar{\boldsymbol{\chi}}_{k}^i \\
                                                                       \boldsymbol{1}_n \\
                                                                      \end{array} \right]
\end{align}
\noindent
where the notation $\bar{\boldsymbol{\chi}}_{k}^i$, $\bar{\boldsymbol{\chi}}_{k+1}^i$ describes the subset of sigma-points related to the state uncertainty, and not to process noise:

\begin{align}
 \bar{\boldsymbol{\chi}}_{k}^i &= \left[ \chi_k^{i,0},...,\chi_k^{i,2n_{\mathbf{x}}}\right]\nonumber \\
 \bar{\boldsymbol{\chi}}_{k+1}^i &= \left[ \chi_{k+1}^{i,0},...,\chi_{k+1}^{i,2n_{\mathbf{x}}}\right].
\end{align}

For a linear system, Equation~\ref{eq:linearization} can be solved exactly for $\mathbf{A}$ and $\mathbf{b}$. For a non-linear system, the residuals can be calculate by casting the linearization as a least-squares problem.

\begin{align}
 \label{eq_leastsquaresdef}
 \left[\mathbf{A}^*,\mathbf{b}^*\right] = \text{argmin}\left(\left\lVert \bar{\boldsymbol{\chi}}_{k+1}^i - \left[\mathbf{A},\mathbf{b}\right] \left[\begin{array}{l l}
   \bar{\boldsymbol{\chi}}_{k}^i \\
   \boldsymbol{1}_n \\
  \end{array} \right]\right\rVert\right)
\end{align}

\noindent
The residual linearization error ($e_{\text{res}}$), defined as: 

\begin{align}
 \label{eq:residual_ls_error}
 e_{\text{res}} &= \text{min}\left(\left\lVert \bar{\boldsymbol{\chi}}_{k+1}^i - \left[\mathbf{A},\mathbf{b}\right] \left[\begin{array}{l l}
   \bar{\boldsymbol{\chi}}_{k}^i \\
   \boldsymbol{1}_n \\
  \end{array} \right]\right\rVert\right) \nonumber \\
 &= \left\lVert \bar{\boldsymbol{\chi}}_{k+1}^i - \left[\mathbf{A}^*,\mathbf{b}^*\right] \left[\begin{array}{l l}
   \bar{\boldsymbol{\chi}}_{k}^i \\
   \boldsymbol{1}_n \\
  \end{array} \right]\right\rVert .
\end{align}
\noindent
The residual error is a direct metric of how well the optimal linear model explains the propagation of sigma points through the nonlinear dynamics. If the underlying model is linear, the residual error is zero. For nonlinear systems, the residual linearization error indicates how locally linear or non-linear the underlying model is in the neighborhood of the sigma-points.

An L-Q factorization can be used to compute $e_{\text{res}}$:

\begin{align}
 \mathbf{L}\mathbf{Q} &= \left[\begin{array}{l l}
                               \bar{\boldsymbol{\chi}}_{k}^i \\
                               \boldsymbol{1}_n \\
                              \end{array} \right], \ \mathbf{L} \in \mathbb{R}^{n_{\mathbf{x}} + 1,  2\cdot n_{\mathbf{x}} + 1} \text{ and } \mathbf{Q} \in \mathbb{R}^{2\cdot n_{\mathbf{x}} + 1,2\cdot n_{\mathbf{x}} + 1} \nonumber
\end{align}
\noindent
where
\begin{align}
 \mathbf{L} &= \left[\mathbf{L}_0, \boldsymbol{0} \right], \quad \mathbf{L}_0 \in \mathbb{R}^{n_{\mathbf{x}} + 1,  n_{\mathbf{x}} + 1} \text{ and } \boldsymbol{0} \in \mathbb{R}^{n_{\mathbf{x}} + 1, n_{\mathbf{x}}}  \nonumber
\end{align}
\noindent
such that $\mathbf{L}_0$ is lower triangular, and therefore cheaply invertible, and $\mathbf{Q}$ is orthonormal:
\begin{align}
 \label{eq_qr_defs}
 \mathbf{Q}\mathbf{Q}^T &= \mathbf{I}.
\end{align}

\noindent
Substituting the factorization from Equation~\ref{eq_qr_defs} into Equation~\ref{eq_leastsquaresdef} yields:

\begin{align}
\label{eq:ls_xform}
\left[\mathbf{A}^*,\mathbf{b}^*\right] &= \text{argmin}\left(\lVert\bar{\boldsymbol{\chi}}_{k+1}^i-\left[\mathbf{A},\mathbf{b}\right]\mathbf{L}\mathbf{Q}\rVert\right) \nonumber \\
&= \text{argmin}\left(\lVert\bar{\boldsymbol{\chi}}_{k+1}^i\mathbf{Q}^T-\left[\mathbf{A},\mathbf{b}\right]\mathbf{L}\rVert\right).
\end{align}

\noindent
Because of the structure of $\mathbf{L}$, the arguments $\left[\mathbf{A},\mathbf{b}\right]$ only effect the first $n_{\mathbf{x}}+1$ columns of the norm in Equation~\ref{eq:ls_xform}. The matrix $\bar{\boldsymbol{\chi}}_{k+1}^i\mathbf{Q}^T$ is partitioned according to what is explained by the linear model ($\hat{\boldsymbol{\chi}}_{k+1}^i$) and what is not explained by a linear model ($\hat{\boldsymbol{\chi}}_{k+1,\text{res}}^i$). In block form, this is written:

\begin{align}
\label{eq:ls_blk_xform}
\left[\mathbf{A}^*,\mathbf{b}^*\right] &= \text{argmin}\left(\left\lVert\left[\hat{\boldsymbol{\chi}}_{k+1}^i, \hat{\boldsymbol{\chi}}_{k+1,\text{res}}^i \right]-\left[\mathbf{A},\mathbf{b}\right]\left[\mathbf{L}_0, \boldsymbol{0} \right]\right\rVert\right).
\end{align}

\noindent
The optimal arguments $\left[\mathbf{A}^*,\mathbf{b}^*\right]$ can be extracted from the partitioned matrix in Equation~\ref{eq:ls_blk_xform} and $\mathbf{L}_0$:

\begin{align}
 \label{eq:leastsquarssoln}
 \left[\mathbf{A}^*,\mathbf{b}^*\right] &= \hat{\boldsymbol{\chi}}_{k+1}^i \mathbf{L}_0^{-1}.
\end{align}

\noindent
The desired linearization error $e_{\text{res}}$, defined in Equation~\ref{eq:residual_ls_error}, simplifies to:

\begin{align}
 \label{eq:ls_err_soln}
 e_{\text{res}} &= \left\lVert \boldsymbol{0}, \hat{\boldsymbol{\chi}}_{k+1,\text{residual}}^i \right\rVert.
\end{align}

\noindent
Note that the linearization error $e_{\text{res}}$ can be calculated without requiring the computed optimal arguments $\mathbf{A}^*$ and $\mathbf{b}^*$ in Equation~\ref{eq:leastsquarssoln}. The scaler residual error in Equation~\ref{eq:ls_err_soln} quantifies how well the temporal propagation of the sigma-points can be explained by a linear model, and therefore how well the local linearity assumptions made by the SP transform hold for the mixand being propagated. This metric is a refinement of the metric proposed by the authors in \citet{havlak2010discrete}, and similar to the metric proposed in \citet{huberadaptive}. Other accuracy metrics for the sigma-point transform have been developed, for example \citet{van2004sigma} proposes a metric based on the Taylor series expansion of the dynamics function, which requires a differentiable dynamics model. If the error metric is above some defined threshold ($e_{\text{res},\text{max}}$), the mixand can be targeted for covariance reduction. However, this metric is 
scalar, and in order to effectively reduce the covariance of the mixand to a specified level of linearization error, it is desirable to identify which sigma-points are explained the least by the optimal linear model. 

Defining $\boldsymbol{\mathcal{E}}_{k+1}^i$ as the difference between the sigma-points propagated through the system model and the optimal linear model,

\begin{align}
 \label{eq:ls_sp_err_def}
 \boldsymbol{\mathcal{E}}_{k+1}^i &= \bar{\boldsymbol{\chi}}_{k+1}^i - \left(\mathbf{A}^*\bar{\boldsymbol{\chi}}_{k}^i + \mathbf{b}^* \right) = \left[\mathcal{E}_{k+1}^{i,1},...,\mathcal{E}_{k+1}^{i,2\cdot n_{\mathbf{x}} + 1}\right].
\end{align}

\noindent
The $j^{\text{th}}$ column of $\boldsymbol{\mathcal{E}}_{k+1}^i$, $\mathcal{E}_{k+1}^{i,j}$ is the residual linearization error associated with the $j^{\text{th}}$ sigma-point. Similar to the linearization error $e_{\text{res}}$,  $\boldsymbol{\mathcal{E}}_{k+1}^i$ can be computed without computing the optimal linearization, given by:

\begin{align}
 \label{eq:ls_err_xform}
 \boldsymbol{\mathcal{E}}_{k+1}^i &= \left[ \boldsymbol{0}, \hat{\boldsymbol{\chi}}_{k+1,\text{residual}}^i \right] \mathbf{Q}.
\end{align}

To effectively reduce the covariance of the mixand in order to reduce the linearization error, the direction along which the linearization error is greatest is identified as the optimal splitting axis ($e_{\text{split}}$). To find $e_{\text{split}}$, the sigma points before propagation ($\chi_k^{i,j}$) are weighted by the norm of their associated residual linearization errors ($\left\lVert \mathcal{E}_{k+1}^{i,j} \right\rVert$). The first eigenvector of the second moment of this set of weighted sigma points is the axis along which the residual linearization error is the greatest, and is used as the optimal splitting axis.

\subsection{Gaussian Mixand Splitting}
\label{sec:split}
Given the sigma point residual error of the propagated mixand in Equation~\ref{eq:ls_err_xform}, the next step is to adapt the mixture to decrease this error (and increase the accuracy of the temporal propagation). This is accomplished by identifying mixands that are poorly propagated by the SP transform (using Equation~\ref{eq:ls_err_xform}) and replacing them with several mixands with smaller covariances.

The approximation of a GMM by a GMM with fewer mixands has been explored in the literature \cite{huber2008progressive,salmon_reduction,garcia2010levels,hennig2010methods,runnalls2007kullback}, as classical GMM filters experience geometric growth in the number of mixands and require a mixture reduction step in order to remain computationally tractable. The problem here, however, is the reverse -- splitting a mixand in order to reduce the variance of mixands. This problem is only beginning to receive attention in the literature \cite{huberadaptive,havlak2010discrete,ali2007gaussian,psiaki2010gaussian}, although there has been interest in the topic in the machine learning community \cite{SMEM_Ueda_1999,faubel2009split,GMM_EM_Split_zhang}. Other adaptive GMM filters include \citet{caputi1993modified}, which relies on non-Gaussian process noise to increase the number of mixands, and \citet{terejanu2008uncertainty}, which adapts the GMM to nonlinearities by developing an update rule for the mixand weights rather 
than 
by increasing 
the number of mixands.

There are two competing goals in approximating a single Gaussian by a GMM. First, the GMM must closely approximate the original Gaussian. Second, the covariance of the mixands in the GMM must be significantly smaller than the covariance of the original Gaussian. A common metric used to evaluate how closely two distributions approximate each other (e.g. comparing the GMM to the original Gaussian) is the Kullback Leibler divergence (KLD), which measures the relative entropy between two distributions\cite{kullback1968information}. However, the KLD between a Gaussian and a GMM can not be evaluated analytically, and effective approximations can be resource intensive\cite{goldberger2005distance,hershey2007approximating}. An alternative metric is the integral-squared difference (ISD), which has a closed-form analytic solution when comparing a single Gaussian to a Gaussian mixture \cite{williams2003cost}. Because the ISD can be exactly and quickly evaluated, it is used here as the statistical error metric when 
comparing a Gaussian and its GMM \cite{
williams2003cost}.

\begin{align}
 \label{eq:splitting_ISD}
 J_{\text{ISD}} &= \int_x\left(\mathcal{N}(x\vert\mu,\Sigma) - \hat{p}(x)\right)^2dx
\end{align}

Formally, the Gaussian mixand splitting problem is defined as the approximation of a single Gaussian mixand $\mathcal{N}(x \vert \mu,\Sigma)$ by a GMM $\hat{p}(x) = \sum_{i=1}^{N}w_i\cdot\mathcal{N}(x\vert\mu_i,\Sigma_i)$ such that the covariances of the mixands ($\Sigma_i$) are smaller than the original ($\Sigma$), and the ISD between the original Gaussian and its GMM (Equation~\ref{eq:splitting_ISD}) is minimized. The formal problem is to solve for $w_i, \mu_i, \text{ and } \Sigma_i$ for all $i$ such that:

\begin{multline}
 \label{eq:optimization}
 \left[w_i, \mu_i, \Sigma_i\right]_{i=1 ... N} = \\
\text{argmin}\left(\int_x\left(\mathcal{N}(x\vert\mu,\Sigma) - \sum_{i=1}^{N}w_i\cdot\mathcal{N}(x\vert\mu_i,\Sigma_i)\right)^2dx\right).
\end{multline}

Because this optimization is computationally expensive, an off-line optimal solution is computed and stored for use as needed by the anticipation algorithm in real time. Pre-computing an optimal split for real time use requires that the pre-computed split can be applied to an arbitrary single Gaussian $\mathcal{N}(x \vert \mu,\Sigma)$ and splitting axis $\mathbf{e}_{\text{split}}$ (i.e. the general problem). Any single Gaussian and splitting axis combination can be transformed to the problem of splitting a unit, zero-mean Gaussian along the $x$-axis ($\mathbf{e}_1$) using an affine transformation.

Given a Gaussian mixand $\mathcal{N}(x \vert \mu,\Sigma)$ and a splitting axis $\mathbf{e}_{\text{split}}$, the following transformations are applied in order to arrive at the pre-computed problem: 

\begin{align}
 \hat{x} = \mathbf{R}_{\text{x}}\mathbf{T}^{-1}\left(x - \mu\right) \nonumber
\end{align}
where $\mathbf{T}$ is the matrix square-root of $\Sigma$
\begin{align}
 \Sigma = \mathbf{T}\mathbf{T}^T \nonumber
\end{align}
and $\mathbf{R}_{\text{x}}$ is a rotation matrix, computed such that the final splitting axis aligns with the $x$-axis:
\begin{align}
 \mathbf{e}_1 = \mathbf{R}_{\text{x}}\mathbf{T}^{-1}\mathbf{e}_{\text{split}}. \label{eq:split_transform}
\end{align}

\begin{figure}[!t]
  \centering
    \subfigure[Single Gaussian]{\label{fig:tr1}\includegraphics[scale=.17]{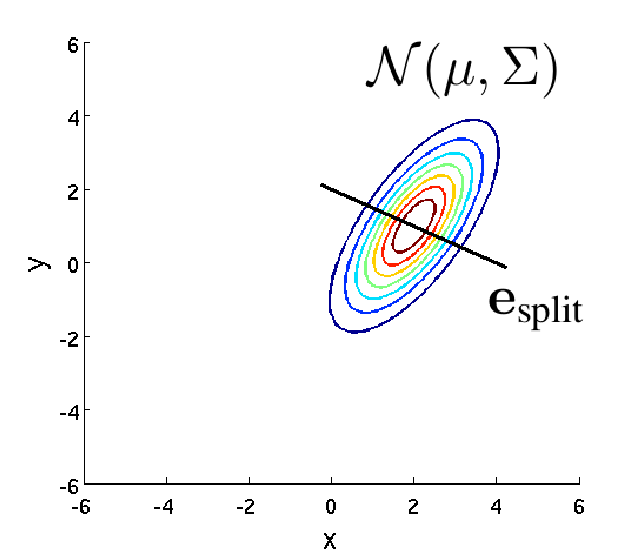}}
    \subfigure[Translated to origin]{\label{fig:tr2}\includegraphics[scale=.17]{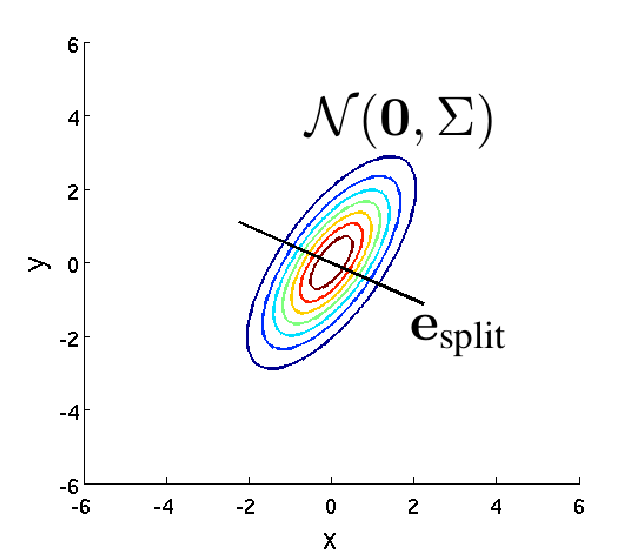}}
    \subfigure[Transformed to unit variance]{\label{fig:tr3}\includegraphics[scale=.17]{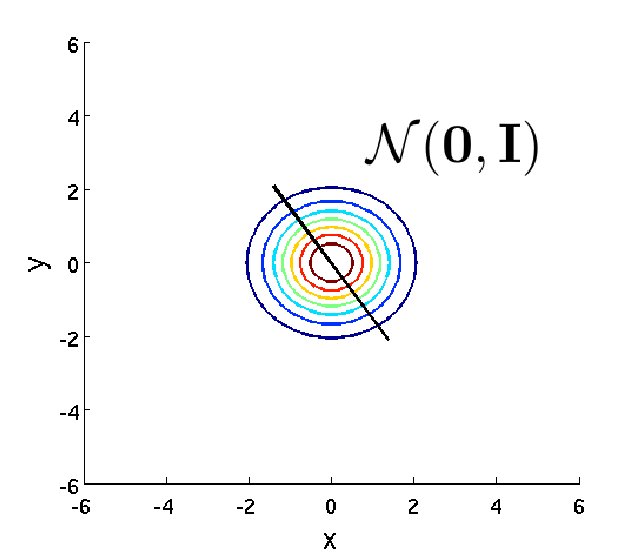}}
    \subfigure[Rotated splitting axis]{\label{fig:tr4}\includegraphics[scale=.17]{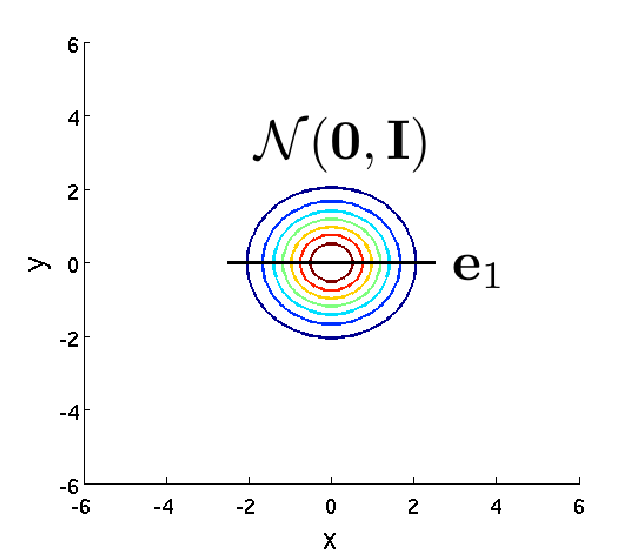}}
  \caption{The transformation from an arbitrary Gaussian/splitting axis to the offline problem}
  \label{fig:split_trans}
\end{figure}

Figure~\ref{fig:split_trans} illustrates the transformation in Equation~\ref{eq:split_transform} applied to a general two dimensional Gaussian. Figure~\ref{fig:tr1} shows the original single Gaussian to be split, and the axis along which the Gaussian is to be split, $e_{\text{split}}$. Figures~\ref{fig:tr2} and \ref{fig:tr3} show the translation of the Gaussian to the origin ($x - \mu$) and the transformation that yields a unit covariance of the single Gaussian. Finally, Figure~\ref{fig:tr4} shows the rotation ($\mathbf{R}_{\text{x}}$) applied that aligns the splitting axis with the $x$-axis.

The splitting problem has now been posed as splitting a zero-mean, unit Gaussian along the $x$-axis into an $N$-component mixture of Gaussians: 

\begin{multline}
 \label{eq:offline_opt}
 \left\{\left(w_1^*, \mu_1^*, \Sigma_1^*\right),...,\left(w_N^*, \mu_N^*, \Sigma_N^*\right)\right\} = \\
 \text{argmin}\left(\int_x\left(\mathcal{N}(x\vert\boldsymbol{0},\mathbf{I}) - \hat{p}(x)\right)^2dx\right).
\end{multline}

\noindent
The solution to this optimization problem is computed off-line, and can be applied at runtime to the general problem (Equation~\ref{eq:optimization}) using the transformation from Equation~\ref{eq:split_transform}:

\begin{align}
 \label{eq:approx_xform}
 w_i &\approx w_i^* \nonumber \\
 \mu_i &\approx \mathbf{T}\mathbf{R}_{\text{x}}^T \cdot \mu_i^* + \mu \nonumber \\
 \Sigma_i &\approx \mathbf{T}\mathbf{R}_{\text{x}}^T \Sigma_i^* \mathbf{R}_{\text{x}} \mathbf{T}^T.
\end{align}

The offline optimization (Equation~\ref{eq:offline_opt}) solves for a Gaussian mixture with an odd number ($N$) elements $\hat{p}(x) = \sum_{i=1}^{N}w_i\cdot\mathcal{N}(x\vert\mu_i,\Sigma_i)$ such that the ISD between the split Gaussian mixture and the Gaussian distribution $\mathcal{N}(x\vert\boldsymbol{0},\mathbf{I})$ is minimized. For an $n_{\mathbf{x}}$ dimensional Gaussian, there are $N\cdot n_{\mathbf{x}}$ parameters associated with the means, $N\cdot\frac{1}{2}(n_{\mathbf{x}}^2+n_{\mathbf{x}})$ parameters associated with the covariances, and $N$ parameters associated with the weights. Due to the large parameter space, the optimization problem is ill-posed, and computationally intractable even for off-line optimization. To address these problems, several constraints are imposed to reduce the parameter space. First, the means $\mu_1$ through $\mu_N$ are restricted to lie on the $x$-axis and be evenly spaced, reducing the optimization parameters associated with the means to a single spread parameter $\
delta_{\mu}$:

\begin{align}
 \mu_i &= \begin{bmatrix}
           \left(i-\frac{N-1}{2}\right) \delta_{\mu} \\
           \boldsymbol{0}
          \end{bmatrix}.
\end{align}

\noindent
This assumption is reasonable, as it spreads the means of split mixture along the splitting axis and only this dimension of the covariance is being targeted for reduction. Furthermore, the derivative of $J_{\text{ISD}}$ with respect to the off-$x$-axis elements of the means evaluated at the proposed $\mu_i$'s is exactly zero.

The next parameter reduction constrains the covariance matrices $\Sigma_1$ through $\Sigma_N$ to be diagonal, equal, and of the form:

\begin{align}
 \Sigma_i &= \begin{bmatrix}
              \sigma & \boldsymbol{0} \\
              \boldsymbol{0}  & \mathbf{I}
             \end{bmatrix}  \nonumber\\
\end{align}
\noindent
where
\begin{align}
 \label{eq:split_var_rest}
 \sigma &\in \left[0,1\right].
\end{align}

\noindent
This targets only the covariance along the splitting axis; all other elements of the split covariance matrices $\sigma$ are optimally $1$ (the derivate of $J_{\text{ISD}}$ with respect to these elements evaluated at the proposed value is zero), so they are not considered. This step reduces the parameters associated with the variance matrices to a single parameter $\sigma$.

Finally, a further reduction to the parameter space is made by recognizing that the optimal weights ($w_1$ through $w_N$) for a given set of means ($\mu_1$ through $\mu_N$) and variances ($\Sigma_1$ through $\Sigma_N$) can be found using a quadratic program, removing the weights from the parameter search space entirely. Expanding the ISD, defined in Equation~\ref{eq:splitting_ISD}, yields:

\begin{align}
 J_{\text{ISD}} &= \int_x \mathcal{N}(x\vert\boldsymbol{0},\mathbf{I})^2 - 2\mathcal{N}(x\vert\boldsymbol{0},\mathbf{I})\hat{p}(x) + \hat{p}(x)^2 dx \nonumber \\
=& \int_x \mathcal{N}(x\vert\boldsymbol{0},\mathbf{I})^2dx - 2\int_x\mathcal{N}(x\vert\boldsymbol{0},\mathbf{I})\hat{p}(x)dx + \int_x\hat{p}(x)^2 dx \nonumber \\
=& J_{1,1} - 2\cdot J_{1,2} + J_{2,2}. \nonumber
\end{align}
\noindent
Williams provides a closed-form solution for each of the above terms:\cite{williams2003cost}
\begin{align}
 J_{1,1} &= \mathcal{N}(\boldsymbol{0}\vert\boldsymbol{0},2\mathbf{I}) \nonumber \\
 J_{1,2} &= \sum_j w_j \cdot \mathcal{N}(\boldsymbol{0}\vert\hat{\mu}_j,\mathbf{I}+\Sigma_j) \nonumber \\
 J_{2,2} &= \sum_i\sum_j w_i w_j \mathcal{N}(\mu_i\vert\mu_j,\Sigma_i+\Sigma_j). \nonumber
\end{align}
\noindent
Re-arranging the terms yields a quadratic program with the weights as arguments:
\begin{align}
 J_{\text{ISD}} &= J_{1,1} -2\mathbf{f}^{T}\mathbf{w} + \mathbf{w}^{T}\mathbf{H}\mathbf{w}, \quad \text{where}\nonumber
\end{align}
\noindent
where
\begin{align}
 \mathbf{w} &= \left[w_1, ... ,w_N\right]^{T} \nonumber \\
 \mathbf{H}_{l,k} &= \mathcal{N}(\mu_l\vert\mu_k,\Sigma_l+\Sigma_k) \nonumber \\
 f_l &= \mathcal{N}(\boldsymbol{0}\vert\mu_l,\mathbf{I}+\Sigma_l) \nonumber
\end{align}

\noindent
such that
\begin{align}
 \label{eq:quad_prog_isd}
 w_i \geq 0 \quad \forall i \nonumber \\
 \sum_i w_i = 1.
\end{align}

\noindent
Equation~\ref{eq:quad_prog_isd} can be solved with a constrained quadratic program to find the optimal weights for the split, $w_i^*$. This allows the weights to be removed from the parameter space, as the optimal weights can be easily computed for each set of parameters considered.

The resulting parameter search space includes only the spacing of the means ($\delta_{\mu}$) and the reduction of the covariance ($\sigma$), making a high-resolution exhaustive parameter search possible. The optimization is an interesting trade between goals. First, $J_{\text{ISD}}$ should be small, as this represents the error introduced by approximating the original Gaussian by the split distribution. Second, $\sigma$ should be small, as this reduces the effects of nonlinearities in the propagation. Finally, $N$ should be small in order to create manageable computation requirements.

\begin{figure}[!t]
 \centering
  \includegraphics[scale=.3]{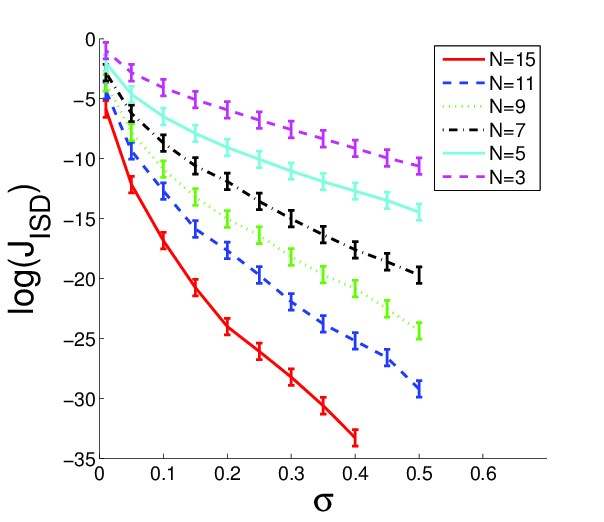}
 \caption{$J_{\text{ISD}}$ as a function of $N$ and $\sigma$, collected for $5000$ randomly generated two-dimensional Gaussian distributions and splitting axes. The one-sigma error bars are shown}
 \label{fig:ISD_N_sigma}
\end{figure}

Figure~\ref{fig:ISD_N_sigma} plots the mean and standard deviation of the ISD as a function of $\sigma$, and $N$ for $5000$ randomly generated examples of two-dimensional Gaussians distributions and splitting axes. For each value of $N$, $\sigma$ is varied between $0.01$ and $0.5$ and the optimal spread parameter and weights are computed. The optimized split is then applied to $5000$ randomly generated two-dimensional Gaussians and splitting axes, and the ISD between each Gaussian and the resulting GMM after the split is computed. Note that the tight error bounds indicate that the optimized split consistently works well, even as Gaussian/splitting axis pairs vary, supporting the approximation in Equation~\ref{eq:approx_xform}. Intuitively, $\sigma = 1$ minimizes the ISD, while $\sigma = 0$ most reduces the variance along the x-axis. It is proposed here to first choose a single value for $\sigma$, and then compute the optimal values for the spread parameter $\delta_{\mu}$ and the weights $w_i$ for different 
values of $N$.

\subsection{Benchmark Splitting Examples}
\label{sec:split_examples}

To explore the impact of the parameter choices $N$ and $\sigma$ on the accuracy of propagation of a single Gaussian through a nonlinear temporal model, two nonlinear models are used as benchmark examples. The first is the univariate non-stationary growth model (UNGM):\cite{kotecha2003gaussian,doucet2000sequential}

\begin{align}
 \label{eq:ungm}
 x_{k+1} = \alpha x_{k} + \beta \frac{x_k}{1+x_k^2} + \gamma \cos{1.2k} \nonumber \\
 \alpha = 0.3, \beta = 1, \gamma = 1
\end{align}

\noindent
and the second is a univariate cubic:

\begin{align}
 \label{eq:cubic}
 x_{k+1} = a x_k^3 + b x_k ^2 + c x_k + d \nonumber \\
 a = 6, b = 1, c = 1, d = 1.
\end{align}

\noindent
The UNGM is commonly used to evaluate the performance of nonlinear filters\cite{kotecha2003gaussian,doucet2000sequential}. Also, for both the UNGM and the cubic model, the propagated distribution $p(x_{k+1})$ can be found analytically for comparison.

The accuracy of the GMM splitting routine is evaluated using 100 randomly generated samples; each is a normal distribution (means uniformly distributed between -2 and 2, and variances uniformly distributed between 0 and 2) and propagated through the UNGM (Equation~\ref{eq:ungm}) and cubic (Equation~\ref{eq:cubic}) models using cached splits. In order to explore the sensitivities in the proposed anticipation algorithm, the cached splits are generated using different variance reductions ($\sigma$) and different numbers of mixands in the split ($N$). The GMM approximation to the propagated distribution $\hat{p}(x_{k+1})$ is compared to the exact propagated distribution $p(x_{k+1})$. The performance metric used to evaluate the accuracy of propagation is the numerically evaluated KLD between the propagated mixture and the true distribution, or:

\begin{align}
 \label{eq:comp_kld}
 \text{KLD} = \int_{x_{k+1}}\log\left(\frac{\hat{p}(x_{k+1})}{p(x_{k+1})}\right)\hat{p}(x_{k+1})dx_{k+1}.
\end{align}

\begin{figure}[!t]
 \centering
  \subfigure[\label{subfig:ungm_klds}]{\includegraphics[scale=.17]{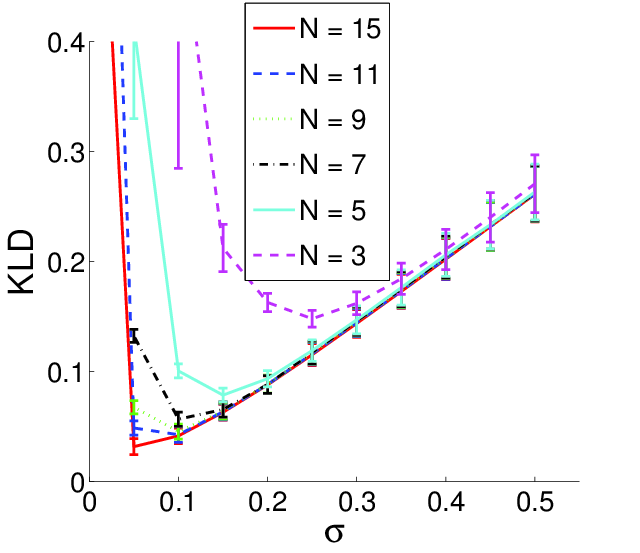}}
  \subfigure[\label{subfig:cubic_klds}]{\includegraphics[scale=.17]{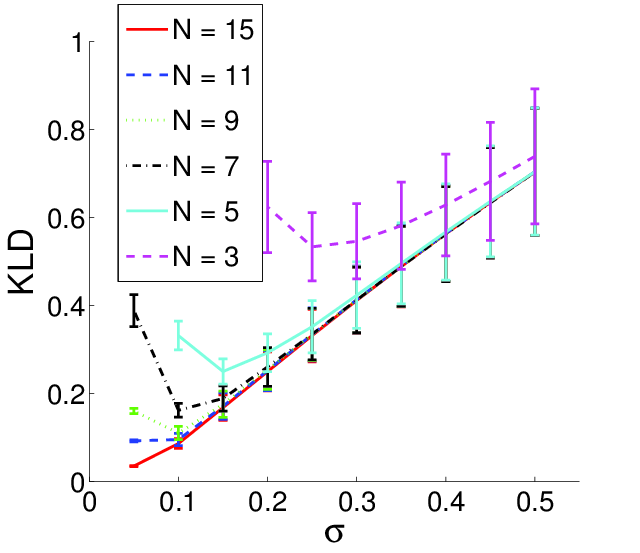}}
 \caption{Accuracy of propagation for the UNGM model \ref{subfig:ungm_klds} and cubic model \ref{subfig:cubic_klds}. For comparison, the mean and variance of the computed KLDs between a propagation with no splitting and the truth are $0.5977$ and $0.0630$ (UNGM model) and $1.4918$ and $0.4259$ (cubic model).}
 \label{fig:klds_ungm_cubic}
\end{figure}

Figure~\ref{fig:klds_ungm_cubic} plots the mean and standard deviation of the KLD between the true distribution $p(x_{k+1})$ and the GMM approximation $\hat{p}(x_{k+1})$ as a function of $N$ and $\sigma$. The trends in each are similar. As $\sigma$ decreases from $0.5$, the accuracy of the split mixtures increases (smaller KLD compared to the true propagated distribution $p(x_{k+1})$).  As $\sigma$ becomes very small, approaching zero, the split mixture becomes a poor approximation of the prior, introducing error in the propagation (higher KLD values). As expected, larger values of $N$ allow smaller values of $\sigma$, and therefore more accurate mixture propagation. Even the least aggressive splits ($\sigma = 0.5$) result in KLDs of approximately one-half that of propagation with no splitting, while more aggressive splits show KLDs of approximately one-tenth that of the no-split solution.

These example problems can also be used to analyze how effective the linearity metric $e_{\text{res}}$ (Equation~\ref{eq:ls_err_soln}) is at predicting propagation errors. $e_{\text{res}}$ is stored for each single Gaussian propagated through the UNGM and the cubic function, and the Pearson correlation between the residual linearization error, a metric that predicts the error in propagation, and the KLD between the no-split solution and truth, which is the actual measured propagation error, is computed. The Pearson correlation coefficient is $0.778$ (UNGM) and $0.535$ (cubic model), indicating strong correlation between the residual linearization error metric and the actual KLD from the true mixture after propagation. 

These two example problems indicate that the linearity criteria and the mixand splitting algorithms greatly improve the accuracy of propagation of a single Gaussian through a non-linear model. The choice of the linearity threshold ($e_{\text{res},\text{max}}$) and of the split parameters $N$ and $\sigma$ are problem dependent, but can be explored off-line prior to implementation.

\section{Experimental Results}

Two systems, one simulated and one experimental, are used to analyze and evaluate the hGMM anticipation algorithm. In both systems, the hGMM is used to predict the future behavior of a tracked car driving on a known roadmap. Additionally, the case of the 2007 Cornell-MIT autonomous collision is studied anecdotally. In order to manage mixture complexity, the GMM reduction method proposed in \citet{runnalls2007kullback} is used in the following examples.

\subsection{Simulation}
\label{sec:example_sim}
The simulation uses the hGMM to predict the future behavior of a simple simulated car driving in a Manhattan grid. Implementing in simulation allows the assumed dynamics model to exactly match the true dynamics model of the simulated obstacle vehicle. This allows the performance of the hGMM in predicting the distribution over future obstacle vehicle states to be studied independently of the assumed dynamics model.

The hGMM algorithm is implemented using a four-state obstacle vehicle model described in reference\cite{havlak2010discrete}. Obstacle vehicles are assumed to drive on a known road network and to obey specified traffic regulations. Obstacle vehicles are modeled as 4-state bicycle robots, with the continuous state:

\begin{align}
 \label{eq:fourstate}
 \mathbf{x}_k^{\text{C}} = \begin{bmatrix}
                   x_k \\
                   y_k \\
                   v_k \\
                   \theta_k
                  \end{bmatrix}
\end{align}

\noindent
where $x_k$ and $y_k$ are the two-axis position of the center of the rear axle, $v_k$ is the speed of the obstacle vehicle, and $\theta_k$ is the heading of the obstacle vehicle, all at time $k$. The continuous dynamics model for the obstacle vehicle is simple, though non-linear.

\begin{align}
 \label{eq:fourstate_dynamics}
 \mathbf{x}^{\text{C}}_{k+1} = f(\mathbf{x}^{\text{C}}_k, \mathbf{u}_k) = \begin{bmatrix}
                                                  x_k + \Delta t \cdot \cos\left(\theta_k\right)\cdot v_k \\
                                                  y_k + \Delta t \cdot\sin\left(\theta_k\right)\cdot v_k \\
                                                  v_k + \Delta t \cdot (u_{1,k} + \mathbf{v}_{1,k}) \\
                                                  \theta_k + \Delta t \cdot l \cdot v_k \cdot (u_{2,k} + \mathbf{v}_{2,k})
                                                 \end{bmatrix}
\end{align}

\noindent
Control inputs are throttle ($u_{1,k}$) and steering angle ($u_{2,k}$), and zero-mean, Gaussian process noise ($\mathbf{v}\sim\mathcal{N}(\boldsymbol{0},\mathbf{Q})$) enters on the control inputs. The time step is $\Delta t=0.1$ seconds in this example.

The hybrid aspect of the obstacle vehicle anticipation problem enters through an assumed route planning controller, where the discrete states are road segments in the map. The discrete dynamics function assumes that the route planning controller takes all possible routes with equal probability. A path following controller is assumed, which generates control inputs $u_{1,k}$ and $u_{2,k}$ that enables the obstacle vehicle to follow the current route and maintain a speed of $10$ meters per second:

\begin{align}
 \label{eq:sim_control}
 \mathbf{u}_k = \begin{bmatrix}
                 u_{1,k} \\
                 u_{2,k}
                \end{bmatrix} = h(\mathbf{x}_k^{\text{C}}, \mathbf{x}_k^{\text{D}}).
\end{align}

\noindent
This path following controller (Equation~\ref{eq:sim_control}) is composed with the bicycle dynamics (Equation~\ref{eq:fourstate_dynamics}) to arrive at the continuous dynamics function $f^{\text{C}}$ (Equation~\ref{eq:dyn_mdl}) required by the hGMM:

\begin{align}
 \mathbf{x}^{\text{C}}_{k+1} = f^{\text{C}}(\mathbf{x}^{\text{D}}_{k+1},\mathbf{x}^{\text{C}}_{k},\mathbf{v}_k) = f(\mathbf{x}^{\text{C}}_k, h(\mathbf{x}_k^{\text{C}}, \mathbf{x}_k^{\text{D}})) .
\end{align}

The anticipation algorithm is used with a time horizon of $3.5$ seconds (approximately the time required for an obstacle to fully traverse an intersection) to predict the behavior of an obstacle car in several different scenarios. For comparison to the hGMM, a large particle set is used to approximate the true distribution of the obstacle state at future times. Three example scenarios (straight road, turn, and intersection) are used to compare the output of the hGMM algorithm to the approximate true distribution. The negative log-likelihood (NLL) is used to evaluate the difference between the true distribution, as represented by a particle set, and the hGMM approximation. The performance of the hGMM is evaluated for different values of $e_{\text{res},\text{max}}$, and compared to the baseline of the hGMM algorithm with no splitting, which is equivalent to a single Gaussian UKF anticipation algorithm.

\begin{figure*}[!t]
 \centering
  \subfigure[\label{subfig:nll_straight}Straight road]{\includegraphics[scale=.26]{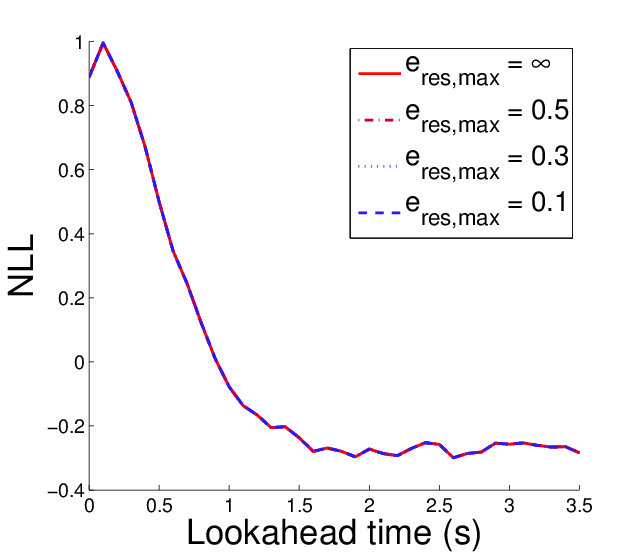}}
  \subfigure[\label{subfig:nll_turn}Approaching a turn]{\includegraphics[scale=.26]{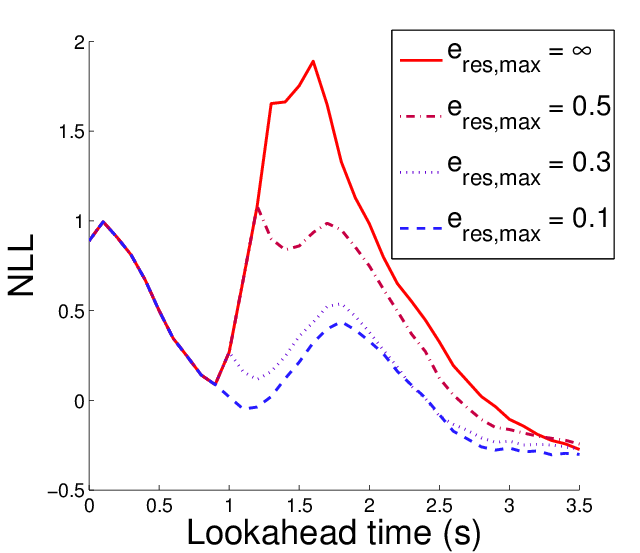}}
  \subfigure[\label{subfig:nll_int}Approaching an intersection]{\includegraphics[scale=.26]{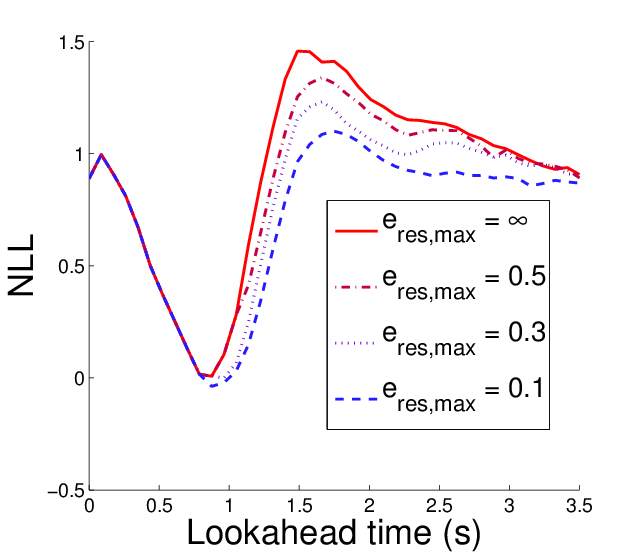}}
 \caption{NLL between ``truth'' particle set and computed probability distributions plotted against the lookahead time (up to a $3.5$ second anticipation horizon) for an obstacle vehicle on a straight road \ref{subfig:nll_straight}, approaching a turn \ref{subfig:nll_turn}, and approaching an intersection \ref{subfig:nll_int}.}
 \label{fig:NLL_car}
\end{figure*}

Figure~\ref{fig:NLL_car} shows the NLL metric for the three considered scenarios. The NLL is plotted as a function of lookahead time. Results are compared for four values of $e_{\text{res},\text{max}}$, for each scenario: the hGMM with $e_{\text{res},\text{max}} = 0.1$, the hGMM with $e_{\text{res},\text{max}} = 0.3$, the hGMM with $e_{\text{res},\text{max}} = 0.5$, and finally the hGMM with $e_{\text{res},\text{max}} = \inf$; the latter is equivalent to a single Gaussian UKF predictor. When the obstacle vehicle is traveling on a straight road (Figure~\ref{subfig:nll_straight}), all of the anticipation predictors show similar performance, because the dynamics model is very close to linear in this case. In Figures~\ref{subfig:nll_turn},\ref{subfig:nll_int}, the hGMM anticipation algorithm shows significant performance improvement over the no splitting predictor, as non-linearities have greater impact on the accuracy of propagation. The peaks in NLL that appear in the second and third scenario correspond to 
the future time when the vehicle is anticipated to be negotiating the bend 
in the road, when the model is exhibiting a high degree of non-linear behavior. Once the turn is negotiated, and the vehicle is anticipated to be on the straight road after the turn, the non-linearities fade and the NLL decreases again.

\begin{figure*}[!t]
 \centering
  \subfigure[\label{subfig:car_int_1}$t_{\text{LA}}=0.1$s]{\includegraphics[scale=.26]{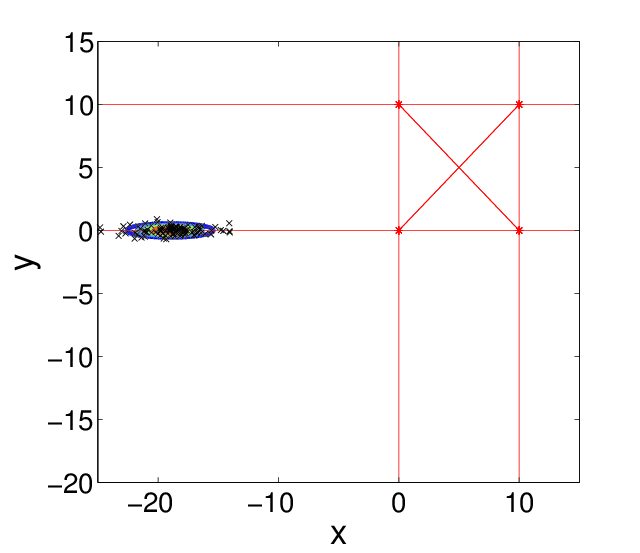}}
  \subfigure[\label{subfig:car_int_2}$t_{\text{LA}}=2$s]{\includegraphics[scale=.26]{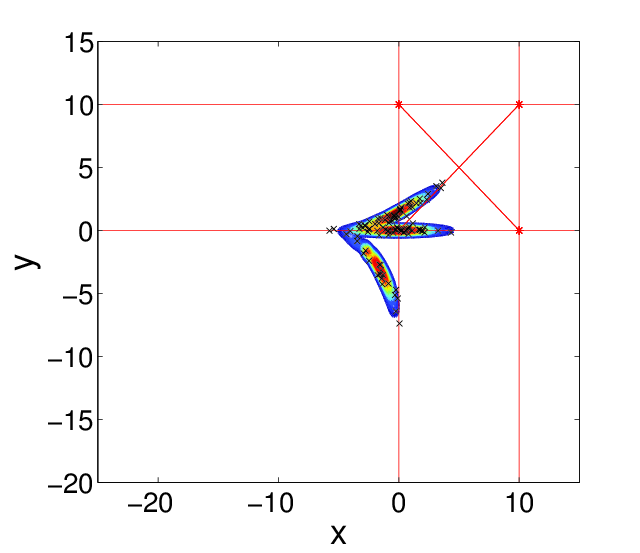}}
  \subfigure[\label{subfig:car_int_3}$t_{\text{LA}}=3$s]{\includegraphics[scale=.26]{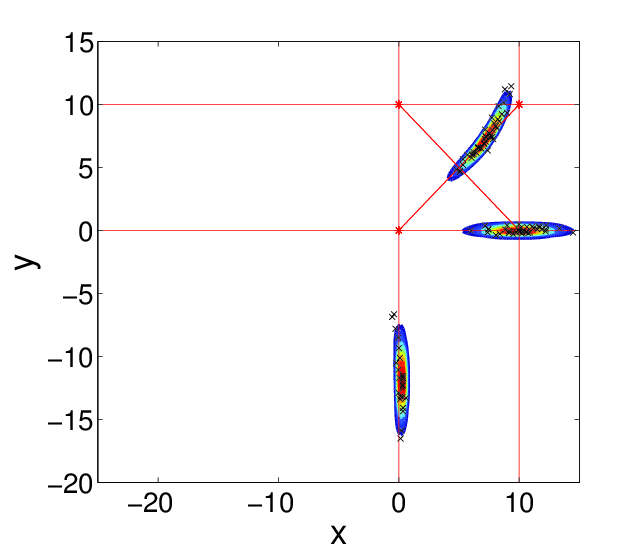}}
 \caption{Anticipated distribution $p(x,y)$ (position of the rear axle) as obstacle traverses an intersection. The time at which the vehicle is observed is $t$, and the time into future the obstacle is anticipated is $t_{\text{LA}}$. The particle set used as truth is overlaid as black dots.}
 \label{fig:car_int}
\end{figure*}

Figure~\ref{fig:car_int} shows the hGMM anticipated distributions of the rear-axle position for the intersection scenario, for $e_{\text{res},\text{max}} = 0.1$. The hGMM anticipation algorithm clearly captures the non-Gaussian nature of the probability masses that turn right and left. This example illustrates both the splitting due to the discrete dynamics (seen by separate probability masses taking each of the three options available at the intersection) and the splitting due to nonlinearities (seen by the non-Gaussian shape of the probability masses making the left and right turns). This figure also demonstrates why Figure~\ref{subfig:nll_int} has a peak just before 2 seconds lookahead -- the non-Gaussian shape of the individual probability masses is at its highest as the obstacle vehicle is anticipated to be negotiating the intersection, as in Figure~\ref{subfig:car_int_2}.


\subsection{Experimental Data}
\label{sec:clif_data}
In order to validate the probabilistic anticipation approach on more realistic problems, the hGMM algorithm was evaluated on a set of tracked vehicle data near intersections. The data set used for this validation is the 2007 Columbus large image format (CLIF 2007) data set made available by the United States Air Force \cite{CLIF2007}. The data set consists of aerial imagery of an urban environment, at approximately two frames per second, collected by a large-format electronics observation platform. A large number of vehicles are observed driving in a variety of conditions.  The validation focuses on vehicles observed traversing intersections, in relatively light traffic, that have the right-of-way. A total of 40 vehicle tracks at three different intersections are used.

In the proposed validation approach, an observed vehicle is tracked at time $t$, giving a state estimate. This state is then predicted forward in time using the hGMM with various values for $e_{\text{res},\text{max}}$, and an anticipation horizon equal to the time over which the vehicle is tracked. The estimate of the observed vehicle state at time $t$ is used as the initial condition for the hGMM; subsequent measurements are used to evaluate how well the hGMM predicts the behavior of the vehicle. The same vehicle behavior model described in Section~\ref{sec:example_sim} (Equation~\ref{eq:fourstate_dynamics}) is used in this experimental validation to describe the dynamics of observed vehicles. Successive measurements of the tracked obstacle at times after $t$ are compared to the anticipated distribution over the vehicle state to evaluate how well the hGMM algorithm anticipates the behavior of the tracked vehicle.

\begin{figure}[!t]
 \centering
  \includegraphics[scale=.42]{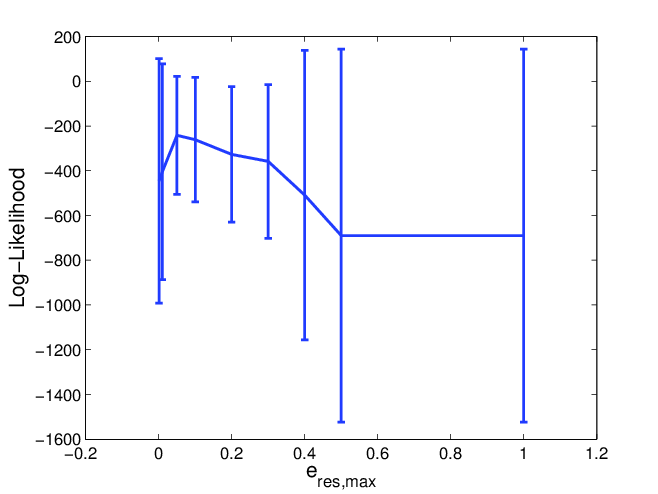}
 \caption{Log-likelihood and standard deviation of observed vehicles given by hGMM anticipation algorithm as a function of $e_{\text{res},\text{max}}$}
 \label{fig:CLIF_NLL}
\end{figure}

Figure~\ref{fig:CLIF_NLL} plots the log-likelihood (LL) of the hGMM anticipation agreeing with CLIF observations as a function of $e_{\text{res},\text{max}}$. At high values of $e_{\text{res},\text{max}}$, the LL is low with larger uncertainty bounds, indicating poorer predictive performance. The figure shows that smaller values for $e_{\text{res},\text{max}}$ provide more accurate anticipation of observed cars, as well as much more consistent performance. At values for $e_{\text{res},\text{max}}$ above approximately $0.5$, the standard deviation becomes quite large. This indicates that the covariance of mixands in the hGMM becomes large enough that the sigma-point propagation becomes a poor and unpredictable approximation. For very small values of $e_{\text{res},\text{max}}$, performance also degrades due to errors introduced by repeated mixand splitting in the hGMM.

\begin{figure}[!t]
 \centering
 \includegraphics[scale=.3]{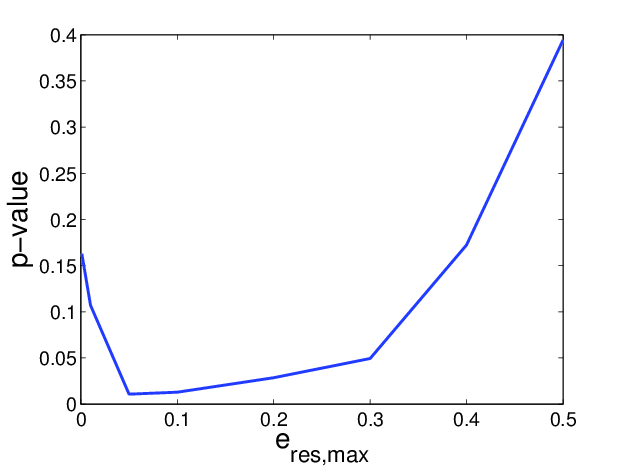}
 \caption{t-test results comparing Log-Likelihood data at different linearity thresholds $e_{\text{res},\text{max}}$ to the data for $e_{\text{res},\text{max}} = 1$.}
 \label{fig:ll_tTest}
\end{figure}

Figure~\ref{fig:ll_tTest} plots the p-values of t-test comparisons between the log-likelihood data for each linearity threshold $e_{\text{res},\text{max}}$ to the log-likelihood data for the maximum linearity threshold ($e_{\text{res},\text{max}} = 1$). This figure shows that the improved performance seen in Figure~\ref{fig:CLIF_NLL} for linearity thresholds $e_{\text{res},\text{max}} = [0.05, 0.1, 0.2]$ is statistically significant, as the corresponding p-values are below $0.05$. For linearity thresholds below this range, the accumulated errors introduced by repeated mixand splitting degrade performance, and the log-likelihood performance is not statistically different from the data taken with a linearity threshold high enough that no splitting occurs ($e_{\text{res},\text{max}} = 1$).

The LL metric studied in Figures~\ref{fig:CLIF_NLL} and \ref{fig:ll_tTest} considers only the likelihood that the observations agree with the anticipated obstacle state, and therefore only evaluates the hGMM at the observations. Because of this, unreasonable predictions (i.e. the car being very far outside the driving lane) are not penalized by this metric. To compliment this metric, a second metric is proposed: the expected off-track error (EOTE). The EOTE is defined as:

\begin{align}
 \label{eq:eote}
 \text{EOTE} = \sum_{k=1}^{K}\int_{x_k} \int_{y_k} d(x_k, y_k)p(x_k, y_k)dy_k dx_k
\end{align}

\noindent
where $K$ is the anticipation horizon, $d(x_k, y_k)$ is the distance of the rear axle position $(x_k, y_k)$ from the lane center, and $\hat{p}(x_k,y_k)$ is the anticipated probability of the vehicle being at that position from the hGMM. This metric is equivalent to the expected value of the out-of-lane position of the observed vehicle. This metric directly measures how reasonable the output of the hGMM is, assuming the observed vehicle is not behaving anomalously.

\begin{figure}[!t]
 \centering
 \includegraphics[scale=.42]{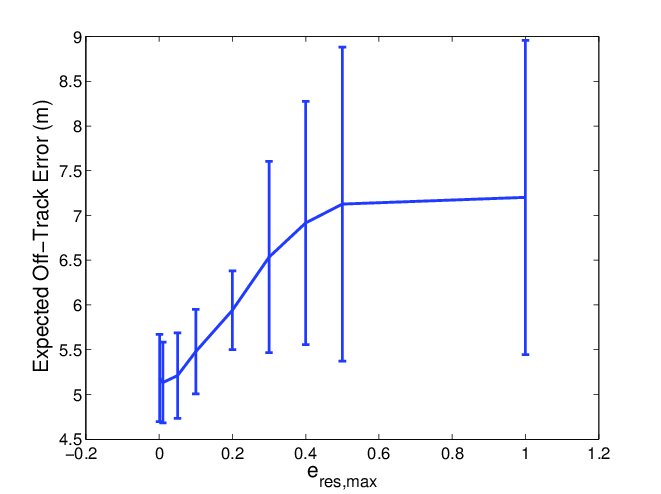}
 \caption{Expected off-track error and standard deviation of observed vehicles given by hGMM anticipation algorithm as a function of the linearity threshold $e_{\text{res},\text{max}}$.}
 \label{fig:eote}
\end{figure}

Figure~\ref{fig:eote} plots the EOTE as a function of the linearity threshold $e_{\text{res},\text{max}}$ used in the hGMM. At high values of $e_{\text{res},\text{max}}$, the EOTE is large and has a large standard deviation, indicating poor anticipation performance. For small values of $e_{\text{res},\text{max}}$, the EOTE and its standard deviation decrease, indicating that the hGMM predictions are more reasonable and more consistent. As with Figure~\ref{fig:CLIF_NLL}, for $e_{\text{res},\text{max}}$ above $0.5$ the standard deviation of this 
metric becomes large as the mixand covariances become too large for the sigma-point approximation.

\begin{figure}[!t]
 \centering
 \includegraphics[scale=.3]{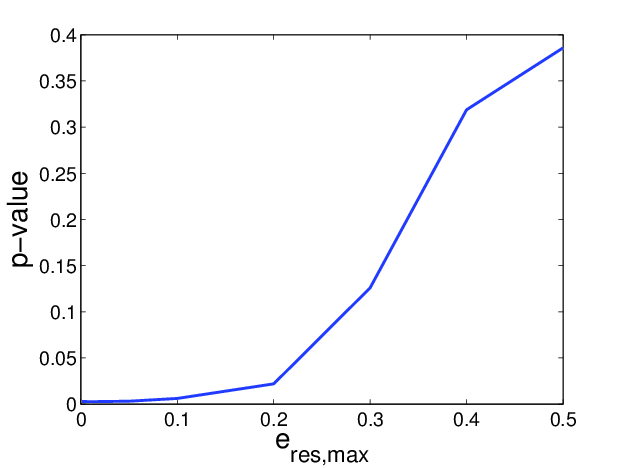}
 \caption{t-test results comparing EOTE data at different linearity thresholds $e_{\text{res},\text{max}}$ to the data for $e_{\text{res},\text{max}} = 1$.}
 \label{fig:eote_tTest}
\end{figure}

Figure~\ref{fig:eote_tTest} plots the results of t-test comparisons between the EOTE data for linearity threshold to the EOTE data for the maximum linearity threshold. These results show that the decreased EOTE seen in Figure~\ref{fig:eote} for small linearity thresholds ($e_{\text{res},\text{max}} \leq 0.2$) is statistically significant.

\begin{figure*}[!t]
 \centering
 \subfigure[\label{subfig:clif_int_1}$t_{\text{LA}}=1$s]{\includegraphics[scale=.21]{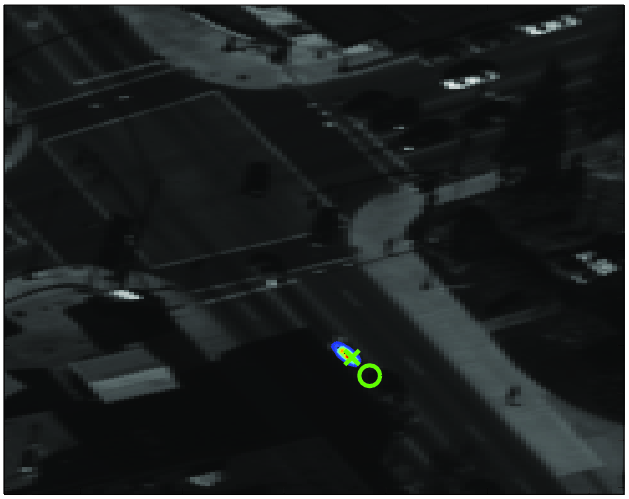}}
 \subfigure[\label{subfig:clif_int_2}$t_{\text{LA}}=3$s]{\includegraphics[scale=.21]{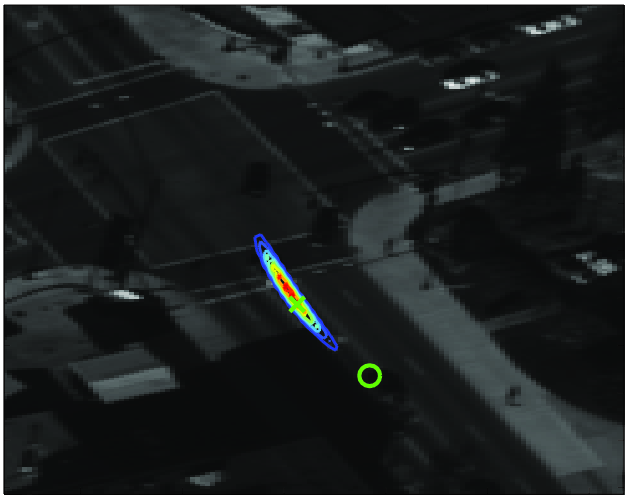}}
 \subfigure[\label{subfig:clif_int_3}$t_{\text{LA}}=7$s]{\includegraphics[scale=.21]{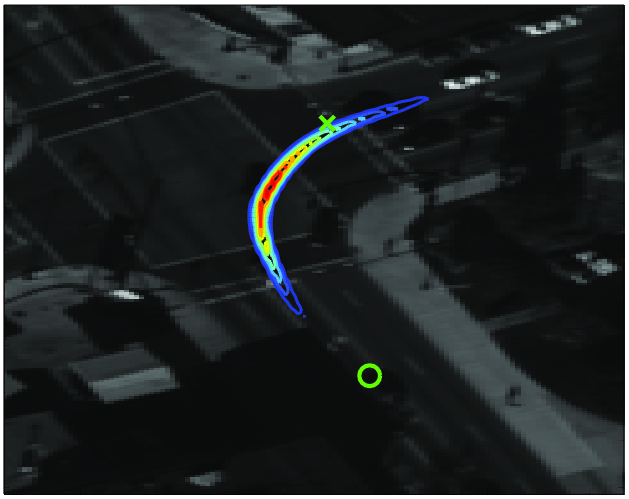}}
 \subfigure[\label{subfig:clif_int_bad_1}$t_{\text{LA}}=1$s]{\includegraphics[scale=.21]{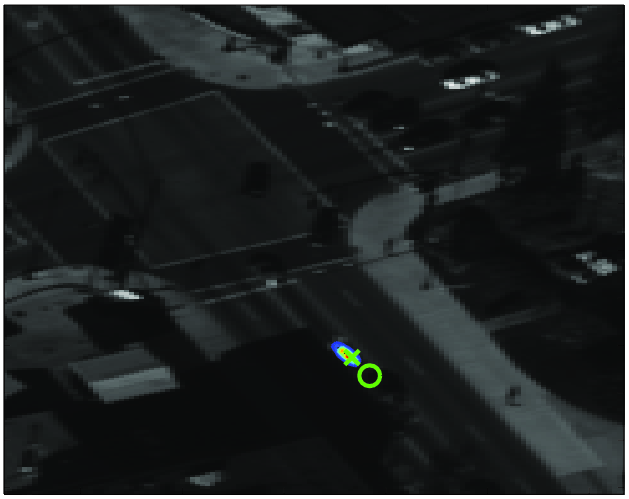}}
 \subfigure[\label{subfig:clif_int_bad_2}$t_{\text{LA}}=3$s]{\includegraphics[scale=.21]{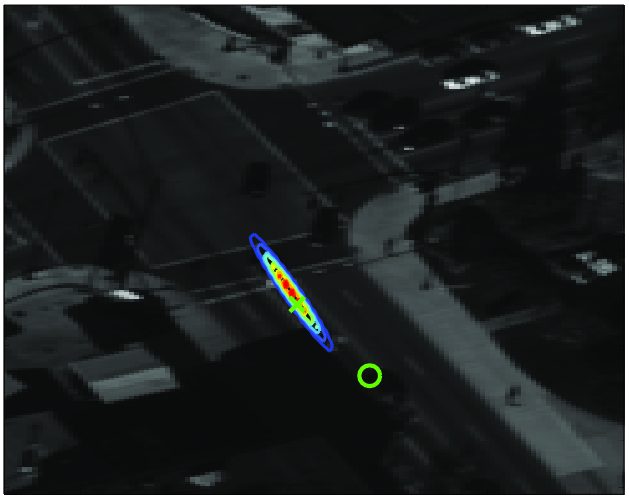}}
 \subfigure[\label{subfig:clif_int_bad_3}$t_{\text{LA}}=7$s]{\includegraphics[scale=.21]{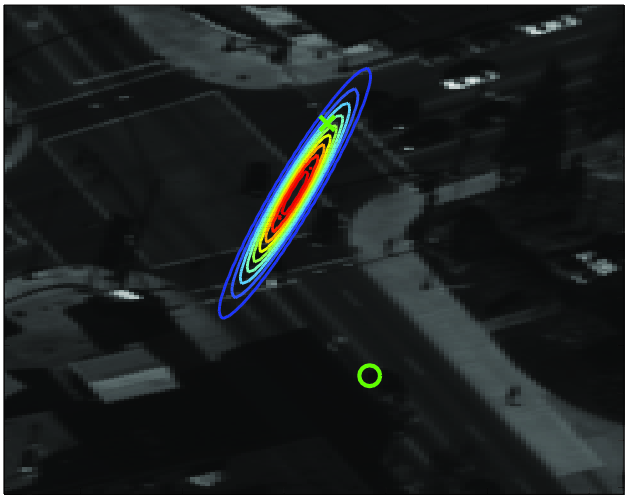}}
 \caption{Anticipation of a real tracked vehicle using the hGMM (\ref{subfig:clif_int_1},\ref{subfig:clif_int_2}, and \ref{subfig:clif_int_3}) and a single Gaussian predictor (\ref{subfig:clif_int_bad_1},\ref{subfig:clif_int_bad_2}, and \ref{subfig:clif_int_bad_3}). The vehicle state at time $t$ is shown as a green circle. The observed vehicle state at the lookahead time is shown as a green cross.}
 \label{fig:clif_int}
\end{figure*}

Figure~\ref{fig:clif_int} shows the results of the hGMM and single Gaussian anticipation algorithms, respectively, for a vehicle making a turn at an intersection. The initial observation of the vehicle (green circle) is plotted along with the anticipated distribution of the rear-axle position at future anticipated times. The actual measurements are plotted as green crosses. Comparing the two approaches, the hGMM predicts the measurements about as well as the single Gaussian method, according to the LL metric in Figure~\ref{fig:CLIF_NLL}. However, Figures~\ref{subfig:clif_int_3} and \ref{subfig:clif_int_bad_3} show that the  hGMM distribution is much more reasonable than the single Gaussian method distribution in predicting locations that are actually on the road network, particularly as the look ahead time increases. This is demonstrated by the large amounts of probability mass outside of the actual expected driving corridor for the vehicle in Figure~\ref{subfig:clif_int_bad_3}, and is quantified by the EOTE 
metric which shows clearly superior performance  of the hGMM algorithm.

\subsection{MIT-Cornell Collision Example}
\label{sec:MITCollision}

\begin{figure}[!t]
 \centering
 \includegraphics[scale=.1]{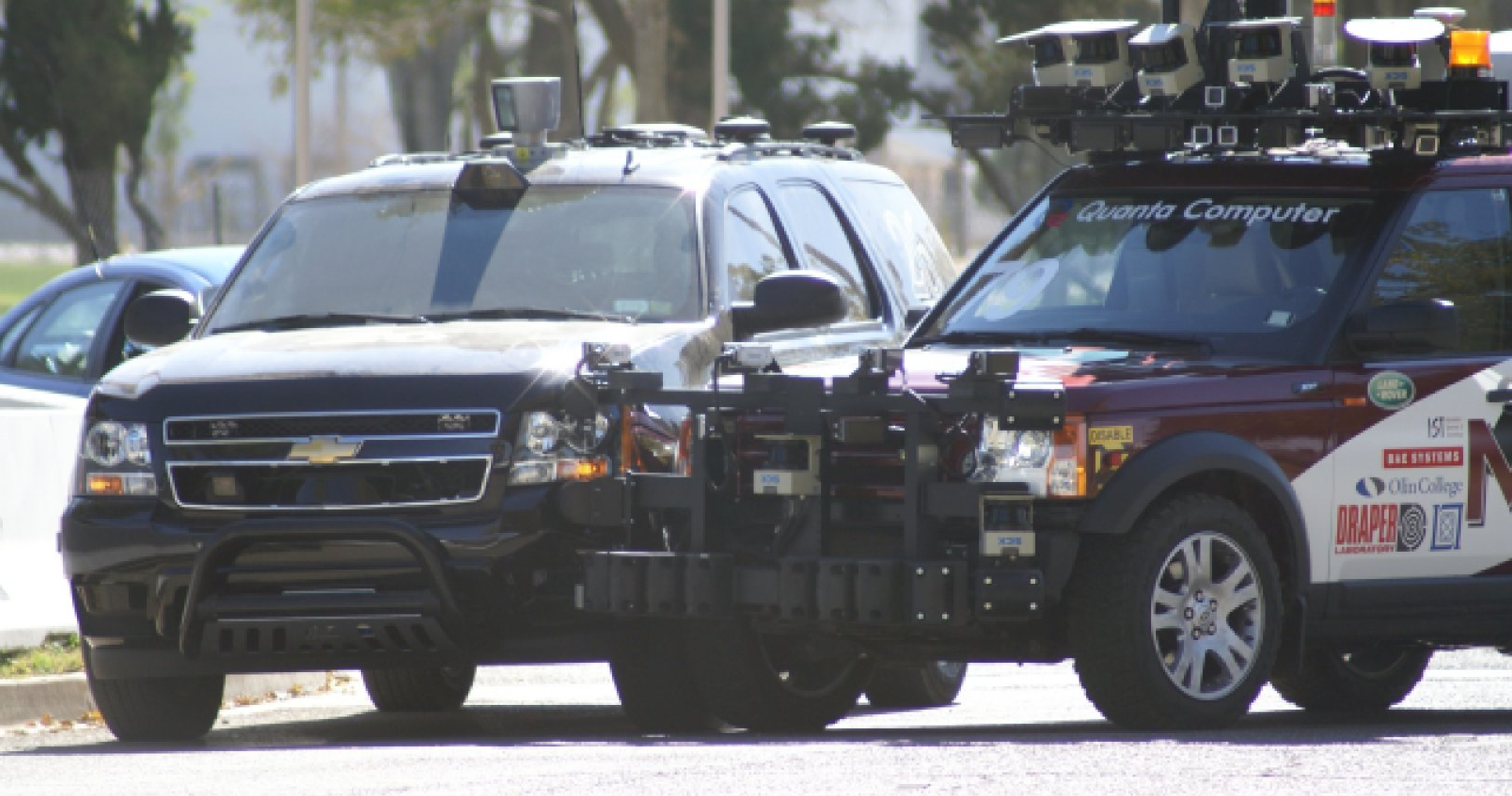}
 \caption{MIT-Cornell collision in the 2007 DUC}
 \label{fig:collisionPhoto}
\end{figure}

A motivating example for the importance of this work is the collision between the Cornell and MIT entries in the 2007 DARPA Urban Challenge (DUC). The collision occurred when the Cornell entry, Skynet, stopped due to a perceived blockage that was the result of a misplaced waypoint in the roadmap. MIT's entry, Talos, observing that Skynet had stopped, initiated a pass. While Talos was executing the pass, Skynet recovered and began moving again. Talos, not recognizing that Skynet was no longer a static obstacle, moved back into the driving lane while Skynet, not able to anticipate the behavior of Talos, continued forward. The result was the low speed collision shown in Figure~\ref{fig:collisionPhoto}. Further details on the collision are available in \citet{fletcher2008cornell}.

To demonstrate the efficacy of the hGMM anticipation algorithm at improving safety, the scenario is re-visited using logged data from the 2007 collision. In this example, Skynet uses the hGMM to anticipate the behavior of Talos (using the simple vehicle model described in~\ref{sec:example_sim}). At the same time, Skynet simulates it's own trajectory for the proposed behavior of resuming motion. The probability of collision between Skynet's proposed motion and the hGMM over Talos' anticipated state is evaluated using the approximation developed in \citet{hardy2010iros}, over a horizon of 3 seconds at 10 Hz.

\begin{figure}[!t]
 \centering
 \subfigure[\label{subfig:colAnt1}$t_{\text{LA}}=0.1$s]{\includegraphics[scale=.17]{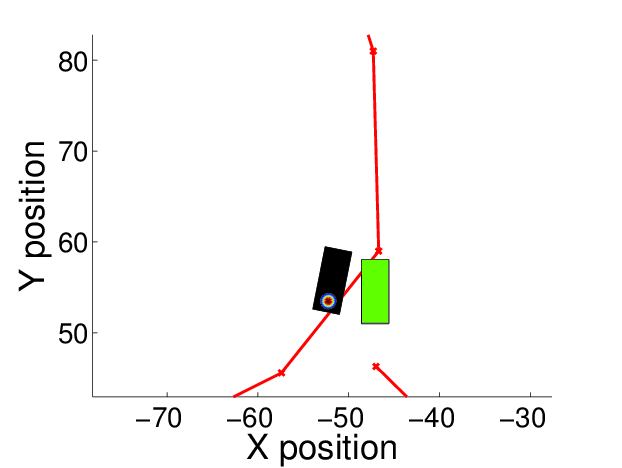}}
 \subfigure[\label{subfig:colAnt2}$t_{\text{LA}}=1.0$s]{\includegraphics[scale=.17]{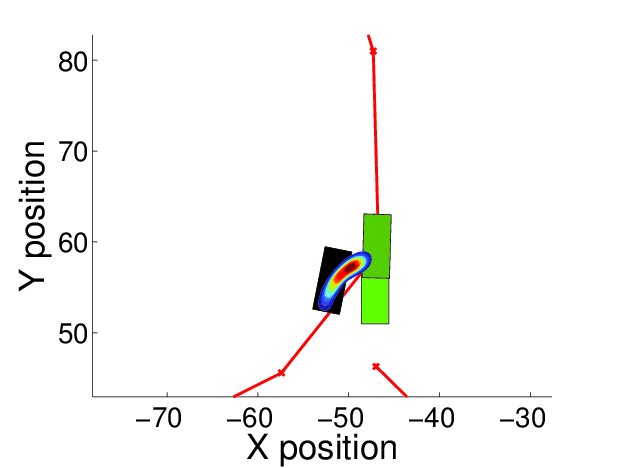}}
 \subfigure[\label{subfig:colAnt3}$t_{\text{LA}}=2.0$s]{\includegraphics[scale=.17]{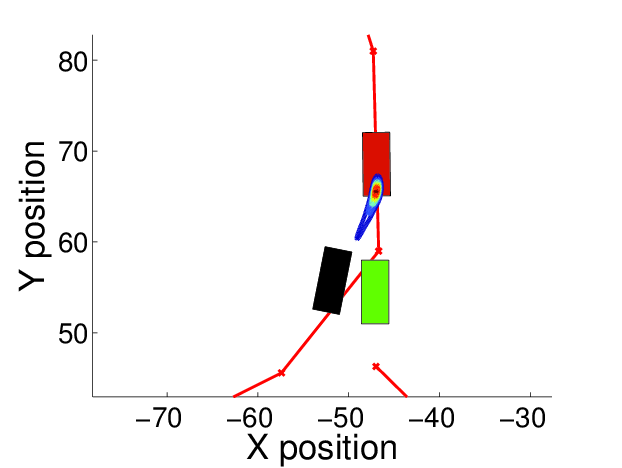}}
 \subfigure[\label{subfig:colAnt4}$t_{\text{LA}}=3.0$s]{\includegraphics[scale=.17]{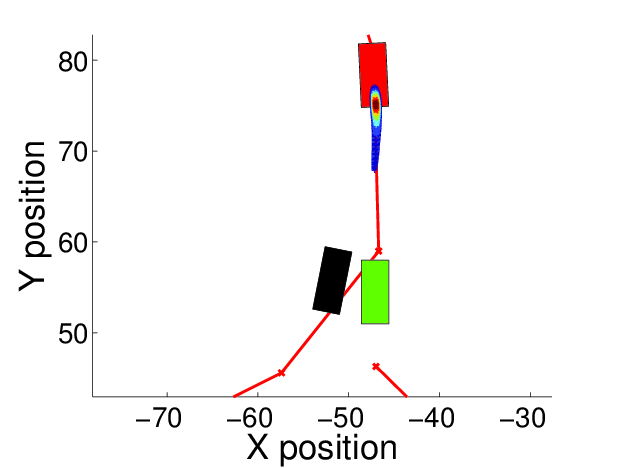}}
 \caption{hGMM anticipation algorithm applied to the MIT-Cornell Collision.}
 \label{fig:colAnt}
\end{figure}

\begin{figure}[!t]
 \centering
 \includegraphics[scale=.3]{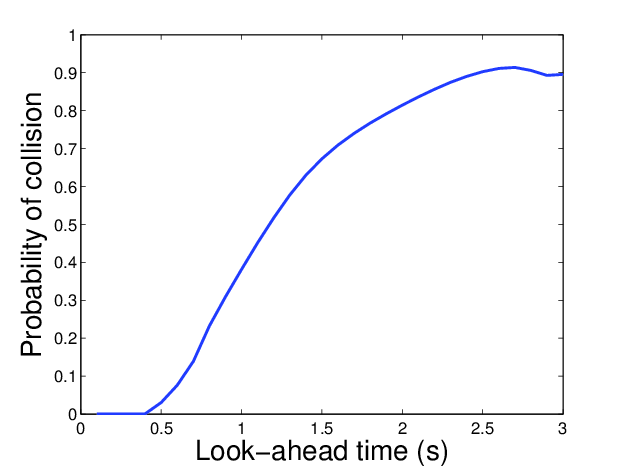}
 \caption{Anticipated probability of collision}
 \label{fig:collProb}
\end{figure}

Figure~\ref{fig:colAnt} shows the results of the anticipation as a function of the lookahead time. Figure~\ref{subfig:colAnt1} shows the initial condition -- Talos is shown as the black car, Cornell is shown as the green car. Figures~\ref{subfig:colAnt2}, \ref{subfig:colAnt3}, and \ref{subfig:colAnt4} show the scenario as it evolves at three different lookahead times. The hGMM probability distribution prediction of Talos is plotted -- in this case, the x-y position of the rear axle. Skynet's proposed position at the given lookahead time is also shown, and the color corresponds to the probability of collision, with green being safe (probability of collision of zero) and red being dangerous (probability of collision of one). Figure~\ref{fig:collProb} plots the anticipated probability of collision as a function of lookahead time. The anticipated probability of collision increases as the anticipation algorithm looks ahead in time, and a collision is almost guaranteed by 2.5 seconds. These figures clearly show 
that even with a basic obstacle model, the hGMM could have predicted, and therefore prevented, the MIT-Cornell collision.

\begin{figure}[!t]
 \centering
 \includegraphics[scale=.35]{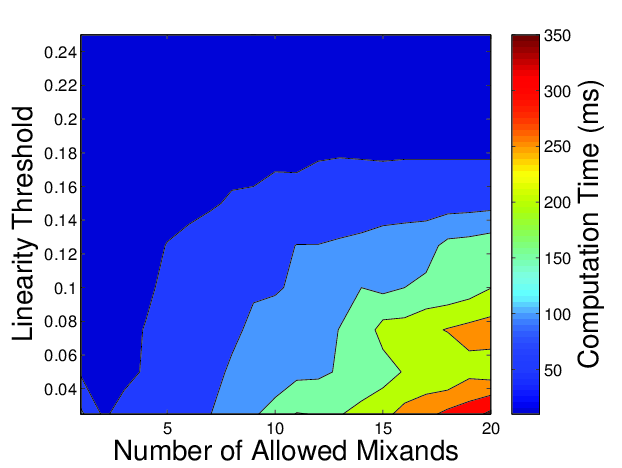}
 \caption{Computation time required to anticipate an obstacle to a 4.5 second horizon as a function of linearity threshold and maximum number of allowed mixands (enforced by Runnall's reduction algorithm) \cite{runnalls2007kullback}.}
 \label{fig:compTime}
\end{figure}

Figure~\ref{fig:compTime} plots the time required to predict an obstacle's state forward 4.5 seconds for a range of values for both the linearity threshold and the maximum number of mixands allowed at the end of each prediction step. The hGMM is implemented in C\# in Skynet's planning algorithms on a modern, Intel Core i7 rackmount computer. The computation time grows with the number of allowed mixands and the inverse of the linearity threshold, as expected.

\begin{figure}[!t]
 \centering
 \includegraphics[scale=.35]{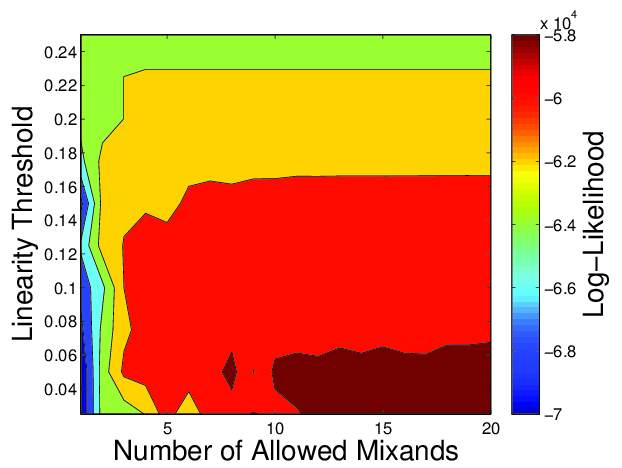}
 \caption{hGMM prediction accuracy as a function of linearity threshold and maximum number of allowed mixands.}
 \label{fig:perfChart}
\end{figure}

Figure~\ref{fig:perfChart} plots the prediction accuracy of the hGMM as a function of the linearity threshold and the maximum number of allowed mixands. Here the prediction accuracy is measured by computing the Log-Likelihood between the hGMM result and a large particle set propagated using the same vehicle dynamics model as the hGMM. Figure~\ref{fig:perfChart} shows that prediction accuracy improves as the linearity threshold becomes smaller, and degrades as the number of allowed mixands becomes small. Note that the performance is independent of the number of allowed mixands until the number of mixands becomes very small. Comparing Figures~\ref{fig:compTime} and \ref{fig:perfChart} provides insight on how to select the linearity threshold and the number of allowed mixands for the assumed dynamics model.

The hGMM is currently implemented in Skynet's planning algorithms with a linearity threshold of $0.1$ and a maximum number of allowed mixands of 10. This implementation allows Skynet to anticipate the behavior of up to three obstacle vehicles to a horizon of 4.5 seconds at greater than 3 Hz. Because the anticipation result is never older than $1/3$ seconds, but extends to 4.5 seconds in the future, this is considered real-time performance.

\section{Conclusion}

An anticipation algorithm is developed which uniquely recognizes and mitigates the impact of non-linearities in the hybrid dynamics function on the accuracy of probability distribution propagation. A unique method for evaluating the accuracy of propagation of a Gaussian distribution through non-linear dynamics is proposed, using the propagated sigma-points from the distribution. In addition, a new method for splitting propagated Gaussian mixands due to nonlinearities is developed. By detecting, and reacting to propagation errors introduced by nonlinearities, the proposed algorithm is shown to have significant improvements in accuracy over standard propagation methods, such as the Unscented Transform.

The behavior of the non-linearity detection and the mixand splitting algorithms are explored using several simulated example problems, and compared to a single Gaussian sigma point method. In the case of benchmark nonlinear problems, the hGMM anticipation algorithm is shown to better approximate the true distributions that arise than the single Gaussian sigma point method. The second set of simulations use an obstacle vehicle driving on a known road network. The accuracy of the anticipated probability distributions is compared across choices for the linearity threshold parameter in the hGMM anticipation algorithm, and compared to a single Gaussian sigma point predictor. Using a large particle set as truth, the hGMM is shown to more accurately capture the behavior of the distribution over future obstacle vehicle states, defined as the negative log-likelihood between the particle set and the anticipated distribution.

The hGMM anticipation algorithm is validated on an experimental data set. Specifically, the hGMM algorithm is used to predict the behavior of vehicles in the CLIF 2007 data set provided by the USAF. The hGMM is compared to a single Gaussian sigma point method by comparing predictions of a tracked vehicle state to observations of the tracked vehicle. The hGMM is shown to provide increased accuracy over the single Gaussian sigma point method using the log-likelihood as a metric. The hGMM is also shown to provide predictions that are more reasonable (i.e. that do not include predictions of anomalous behavior for vehicles that are not behaving anomalously, by the expected off-track error metric) than propagation with no splitting.

Additionally, the hGMM is applied to the scenario leading to the MIT-Cornell collision in the 2007 DUC, and it is shown that the hGMM could have anticipated the collision in time to prevent it.

\bibliographystyle{plainnat}
\bibliography{IEEEabrv,fshavlak}

\begin{thebibliography}{51}
\providecommand{\natexlab}[1]{#1}
\providecommand{\url}[1]{\texttt{#1}}
\expandafter\ifx\csname urlstyle\endcsname\relax
  \providecommand{\doi}[1]{doi: #1}\else
  \providecommand{\doi}{doi: \begingroup \urlstyle{rm}\Url}\fi

\bibitem[Ali-L{\"o}ytty and Sirola(2007)]{ali2007gaussian}
S.~Ali-L{\"o}ytty and N.~Sirola.
\newblock Gaussian mixture filter in hybrid navigation.
\newblock In \emph{Proceedings of The European Navigation Conference GNSS},
  volume 2007, 2007.

\bibitem[Alspach and Sorenson(1972)]{alspach1972nonlinear}
D.~Alspach and HW~Sorenson.
\newblock Nonlinear bayesian estimation using gaussian sum approximations.
\newblock \emph{Automatic Control, IEEE Transactions on}, 17\penalty0
  (4):\penalty0 439--448, 1972.

\bibitem[Bacha et~al.(2008)Bacha, Bauman, Faruque, Fleming, Terwelp, Reinholtz,
  Hong, Wicks, Alberi, Anderson, Cacciola, Currier, Dalton, Farmer, Hurdus,
  Kimmel, King, Taylor, Covern, and Webster]{bacha2008odin}
Andrew Bacha, Cheryl Bauman, Ruel Faruque, Michael Fleming, Chris Terwelp,
  Charles~F. Reinholtz, Dennis Hong, Al~Wicks, Thomas Alberi, David Anderson,
  Stephen Cacciola, Patrick Currier, Aaron Dalton, Jesse Farmer, Jesse Hurdus,
  Shawn Kimmel, Peter King, Andrew Taylor, David~Van Covern, and Mike Webster.
\newblock Odin: Team victortango's entry in the darpa urban challenge.
\newblock \emph{J. Field Robotics}, 25\penalty0 (8):\penalty0 467--492, 2008.

\bibitem[Bishop(2000)]{bishop2000intelligent}
R.~Bishop.
\newblock Intelligent vehicle applications worldwide.
\newblock \emph{Intelligent Systems and Their Applications, IEEE}, 15\penalty0
  (1):\penalty0 78--81, 2000.

\bibitem[Caputi and Moose(1993)]{caputi1993modified}
Mauro~J Caputi and Richard~L Moose.
\newblock A modified gaussian sum approach to estimation of non-gaussian
  signals.
\newblock \emph{Aerospace and Electronic Systems, IEEE Transactions on},
  29\penalty0 (2):\penalty0 446--451, 1993.

\bibitem[Choi et~al.()Choi, Eoh, Kim, Yoon, Park, and Lee]{choi2010analytic}
J.S. Choi, G.~Eoh, J.~Kim, Y.~Yoon, J.~Park, and B.H. Lee.
\newblock Analytic collision anticipation technology considering agents' future
  behavior.
\newblock In \emph{Intelligent Robots and Systems (IROS), 2010 IEEE/RSJ
  International Conference on}, pages 1656--1661. IEEE.

\bibitem[Doucet et~al.(2000)Doucet, Godsill, and Andrieu]{doucet2000sequential}
A.~Doucet, S.~Godsill, and C.~Andrieu.
\newblock On sequential monte carlo sampling methods for bayesian filtering.
\newblock \emph{Statistics and computing}, 10\penalty0 (3):\penalty0 197--208,
  2000.

\bibitem[{Du Toit} and Burdick(2010)]{du-robotic}
N.E. {Du Toit} and J.W. Burdick.
\newblock {Robotic Motion Planning in Dynamic, Cluttered, Uncertain
  Environments}.
\newblock In \emph{2010 IEEE Conference on Robotics and Automation}, pages
  966--973, 2010.

\bibitem[Du~Toit and Burdick(May)]{dutoit2012robot}
N.E. Du~Toit and J.W. Burdick.
\newblock Robotic motion planning in dynamic, cluttered, uncertain
  environments.
\newblock In \emph{Robotics and Automation (ICRA), 2010 IEEE International
  Conference on}, pages 966--973, May.
\newblock \doi{10.1109/ROBOT.2010.5509278}.

\bibitem[Faubel et~al.(2009)Faubel, McDonough, and Klakow]{faubel2009split}
F.~Faubel, J.~McDonough, and D.~Klakow.
\newblock The split and merge unscented gaussian mixture filter.
\newblock \emph{Signal Processing Letters, IEEE}, 16\penalty0 (9):\penalty0
  786--789, 2009.

\bibitem[Ferguson et~al.(2008)Ferguson, Darms, Urmson, and
  Kolski]{ferguson2008detection}
D.~Ferguson, M.~Darms, C.~Urmson, and S.~Kolski.
\newblock {Detection, prediction, and avoidance of dynamic obstacles in urban
  environments}.
\newblock In \emph{2008 IEEE Intelligent Vehicles Symposium}, pages 1149--1154,
  2008.

\bibitem[Fletcher et~al.(2008)Fletcher, Teller, Olson, Moore, Kuwata, How,
  Leonard, Miller, Campbell, Huttenlocher, et~al.]{fletcher2008cornell}
L.~Fletcher, S.~Teller, E.~Olson, D.~Moore, Y.~Kuwata, J.~How, J.~Leonard,
  I.~Miller, M.~Campbell, D.~Huttenlocher, et~al.
\newblock {The mit-cornell collision and why it happened}.
\newblock \emph{Journal of Field Robotics}, 25\penalty0 (10):\penalty0
  775--807, 2008.

\bibitem[Folsom(2011)]{folsom2011social}
T.C. Folsom.
\newblock Social ramifications of autonomous urban land vehicles.
\newblock In \emph{IEEE International Symposium on Technology and Society,
  May}, 2011.

\bibitem[Force(2007)]{CLIF2007}
United States~Air Force.
\newblock 2007 clif data set.
\newblock 2007.
\newblock URL \url{https://www.sdms.afrl.af.mil/index.php?collection=clif2007}.

\bibitem[Fulgenzi et~al.(2009)Fulgenzi, Spalanzani, and
  Laugier]{fulgenzi2009probabilistic}
C.~Fulgenzi, A.~Spalanzani, and C.~Laugier.
\newblock {Probabilistic rapidly-exploring random trees for autonomous
  navigation among moving obstacles}.
\newblock In \emph{Workshop on safe navigation, IEEE International Conference
  on Robotics and Automation (ICRA)}, 2009.

\bibitem[Garcia et~al.(2010)Garcia, Nielsen, and Nock]{garcia2010levels}
V.~Garcia, F.~Nielsen, and R.~Nock.
\newblock Levels of details for gaussian mixture models.
\newblock \emph{Computer Vision--ACCV 2009}, pages 514--525, 2010.

\bibitem[Goldberger and Aronowitz(2005)]{goldberger2005distance}
J.~Goldberger and H.~Aronowitz.
\newblock {A distance measure between gmms based on the unscented transform and
  its application to speaker recognition}.
\newblock In \emph{Ninth European Conference on Speech Communication and
  Technology}. Citeseer, 2005.

\bibitem[Hardy and Campbell(In Press)]{hardy2010iros}
J.~Hardy and M.~Campbell.
\newblock {Contingency Planning over Probabilistic Hybrid Obstacle Predictions
  for Autonomous Road Vehicles}.
\newblock In \emph{IEEE International Conference on Intelligent Robots and
  Systems, 2010}, In Press.

\bibitem[Havlak and Campbell(2010)]{havlak2010discrete}
F.~Havlak and M.~Campbell.
\newblock {Discrete and continuous, probabilistic anticipation for autonomous
  robots in urban environments}.
\newblock In \emph{Proceedings of SPIE}, volume 7833, page 78330H, 2010.

\bibitem[Hennig(2010)]{hennig2010methods}
C.~Hennig.
\newblock Methods for merging gaussian mixture components.
\newblock \emph{Advances in Data Analysis and Classification}, 4\penalty0
  (1):\penalty0 3--34, 2010.

\bibitem[Hershey and Olsen(2007)]{hershey2007approximating}
J.R. Hershey and P.A. Olsen.
\newblock {Approximating the Kullback Leibler divergence between Gaussian
  mixture models}.
\newblock In \emph{IEEE International Conference on Acoustics, Speech and
  Signal Processing, 2007. ICASSP 2007}, volume~4, 2007.

\bibitem[Huber(2011)]{huberadaptive}
M.F. Huber.
\newblock Adaptive gaussian mixture filter based on statistical linearization.
\newblock 2011.

\bibitem[Huber and Hanebeck(2008)]{huber2008progressive}
M.F. Huber and U.D. Hanebeck.
\newblock {Progressive Gaussian mixture reduction}.
\newblock In \emph{Information Fusion, 2008 11th International Conference on},
  pages 1--8. IEEE, 2008.

\bibitem[Huttenlocher et~al.(2008)Huttenlocher, Sergei, Pete, Mike, and
  Fujishima]{huttenlocher2008team}
I.M.M.C.D. Huttenlocher, F.R.K.A.N. Sergei, L.J.C.B.S. Pete, M.N.Z.E.G. Mike,
  and K.H. Fujishima.
\newblock {Team Cornell's Skynet: Robust Perception and Planning in an Urban
  Environment}.
\newblock \emph{Journal of Field Robotics}, 25\penalty0 (8), 2008.

\bibitem[Juan et~al.(2006)Juan, Wu, and McDonald]{juan2006socio}
Z.~Juan, J.~Wu, and M.~McDonald.
\newblock Socio-economic impact assessment of intelligent transport systems.
\newblock \emph{Tsinghua Science \& Technology}, 11\penalty0 (3):\penalty0
  339--350, 2006.

\bibitem[Julier and Uhlmann(1996)]{julier1996general}
S.~Julier and J.K. Uhlmann.
\newblock A general method for approximating nonlinear transformations of
  probability distributions.
\newblock \emph{Robotics Research Group, Department of Engineering Science,
  University of Oxford, Oxford, OC1 3PJ United Kingdom, Tech. Rep}, 1996.

\bibitem[Julier and Uhlmann(1997)]{julier1997new}
S.J. Julier and J.K. Uhlmann.
\newblock A new extension of the kalman filter to nonlinear systems.
\newblock In \emph{Int. Symp. Aerospace/Defense Sensing, Simul. and Controls},
  volume~3, page~26. Spie Bellingham, WA, 1997.

\bibitem[Kotecha and Djuric(2003)]{kotecha2003gaussian}
J.H. Kotecha and P.M. Djuric.
\newblock Gaussian sum particle filtering.
\newblock \emph{Signal Processing, IEEE Transactions on}, 51\penalty0
  (10):\penalty0 2602--2612, 2003.

\bibitem[Kullback(1968)]{kullback1968information}
S.~Kullback.
\newblock \emph{Information theory and statistics}.
\newblock Dover, 1968.

\bibitem[Labs()]{fuberlin_pr}
Autonomous Labs.
\newblock Autonomous car navigates the streets of berlin.

\bibitem[Levinson et~al.(2011)Levinson, Askeland, Becker, Dolson, Held, Kammel,
  Kolter, Langer, Pink, Pratt, et~al.]{levinson2011towards}
J.~Levinson, J.~Askeland, J.~Becker, J.~Dolson, D.~Held, S.~Kammel, J.Z.
  Kolter, D.~Langer, O.~Pink, V.~Pratt, et~al.
\newblock Towards fully autonomous driving: Systems and algorithms.
\newblock In \emph{Intelligent Vehicles Symposium (IV), 2011 IEEE}, pages
  163--168. IEEE, 2011.

\bibitem[Markoff(2010)]{markoff2010google}
John Markoff.
\newblock Google cars drive themselves, in traffic.
\newblock \emph{New York Times}, Oct 2010.

\bibitem[Markoff(2012)]{markoff2012collision}
John Markoff.
\newblock Collision in the making between self-driving cars and how the world
  works, Jan 2012.

\bibitem[Miller et~al.(2008)Miller, Campbell, Huttenlocher, Kline, Nathan,
  Lupashin, Catlin, Schimpf, Moran, Zych, Garcia, Kurdziel, and
  Fujishima]{miller2008team}
Isaac Miller, Mark~E. Campbell, Dan Huttenlocher, Frank-Robert Kline, Aaron
  Nathan, Sergei Lupashin, Jason Catlin, Brian Schimpf, Pete Moran, Noah Zych,
  Ephrahim Garcia, Mike Kurdziel, and Hikaru Fujishima.
\newblock Team cornell's skynet: Robust perception and planning in an urban
  environment.
\newblock \emph{J. Field Robotics}, 25\penalty0 (8):\penalty0 493--527, 2008.

\bibitem[Miura and Shirai(2002)]{miura2002probabilistic}
J.~Miura and Y.~Shirai.
\newblock {Probabilistic uncertainty modeling of obstacle motion for robot
  motion planning}.
\newblock \emph{Journal of Robotics and Mechatronics}, 14\penalty0
  (4):\penalty0 349--356, 2002.

\bibitem[Ohki et~al.(2010)Ohki, Nagatani, and Yoshida]{ohki2010collision}
T.~Ohki, K.~Nagatani, and K.~Yoshida.
\newblock Collision avoidance method for mobile robot considering motion and
  personal spaces of evacuees.
\newblock In \emph{Intelligent Robots and Systems (IROS), 2010 IEEE/RSJ
  International Conference on}, pages 1819--1824. IEEE, 2010.

\bibitem[Petti et~al.(2005)Petti, Fraichard, Rocquencourt, and
  Paris]{Petti05safemotion}
Stephane Petti, Thierry Fraichard, Inria Rocquencourt, and Ecole Des Mines~De
  Paris.
\newblock Safe motion planning in dynamic environments.
\newblock pages 3726--3731, 2005.

\bibitem[Psiaki et~al.(2010)Psiaki, Schoenberg, and Miller]{psiaki2010gaussian}
Mark~L Psiaki, Jonathan~R Schoenberg, and Isaac~T Miller.
\newblock Gaussian mixture approximation by another gaussian mixture for" blob"
  filter re-sampling.
\newblock \emph{AIAA Paper}, \penalty0 (2010-7747):\penalty0 2--5, 2010.

\bibitem[Runnalls(2007)]{runnalls2007kullback}
Andrew~R Runnalls.
\newblock Kullback-leibler approach to gaussian mixture reduction.
\newblock \emph{Aerospace and Electronic Systems, IEEE Transactions on},
  43\penalty0 (3):\penalty0 989--999, 2007.

\bibitem[Salmond(2009)]{salmon_reduction}
D.J. Salmond.
\newblock Mixture reduction algorithms for point and extended object tracking
  in clutter.
\newblock \emph{Aerospace and Electronic Systems, IEEE Transactions on},
  45\penalty0 (2):\penalty0 667--686, April 2009.
\newblock ISSN 0018-9251.
\newblock \doi{10.1109/TAES.2009.5089549}.

\bibitem[Sorenson and Alspach(1971)]{sorenson1971recursive}
H.W. Sorenson and D.L. Alspach.
\newblock Recursive bayesian estimation using gaussian sums.
\newblock \emph{Automatica}, 7\penalty0 (4):\penalty0 465--479, 1971.

\bibitem[Terejanu et~al.(2008)Terejanu, Singla, Singh, and
  Scott]{terejanu2008uncertainty}
Gabriel Terejanu, Puneet Singla, Tarunraj Singh, and Peter~D Scott.
\newblock Uncertainty propagation for nonlinear dynamic systems using gaussian
  mixture models.
\newblock \emph{Journal of guidance, control, and dynamics}, 31\penalty0
  (6):\penalty0 1623, 2008.

\bibitem[Ueda et~al.(1999)Ueda, Nakano, Ghahramani, and Hinton]{SMEM_Ueda_1999}
Naonori Ueda, Ryohei Nakano, Zoubin Ghahramani, and Geoffrey~E. Hinton.
\newblock Smem algorithm for mixture models.
\newblock \emph{NEURAL COMPUTATION}, pages 200--0, 1999.

\bibitem[Uhlmann(1995)]{uhlmann1995dynamic}
J.K. Uhlmann.
\newblock \emph{Dynamic map building and localization: New theoretical
  foundations}.
\newblock PhD thesis, University of Oxford, 1995.

\bibitem[Urmson et~al.(2008)Urmson, Anhalt, Bagnell, Baker, Bittner, Clark,
  Dolan, Duggins, Galatali, Geyer, Gittleman, Harbaugh, Hebert, Howard, Kolski,
  Kelly, Likhachev, McNaughton, Miller, Peterson, Pilnick, Rajkumar, Rybski,
  Salesky, Seo, Singh, Snider, Stentz, Whittaker, Wolkowicki, Ziglar, Bae,
  Brown, Demitrish, Litkouhi, Nickolaou, Sadekar, Zhang, Struble, Taylor,
  Darms, and Ferguson]{urmson2008autonomous}
Chris Urmson, Joshua Anhalt, Drew Bagnell, Christopher~R. Baker, Robert
  Bittner, M.~N. Clark, John~M. Dolan, Dave Duggins, Tugrul Galatali,
  Christopher Geyer, Michele Gittleman, Sam Harbaugh, Martial Hebert, Thomas~M.
  Howard, Sascha Kolski, Alonzo Kelly, Maxim Likhachev, Matthew McNaughton,
  Nick Miller, Kevin Peterson, Brian Pilnick, Raj Rajkumar, Paul~E. Rybski,
  Bryan Salesky, Young-Woo Seo, Sanjiv Singh, Jarrod Snider, Anthony Stentz,
  William Whittaker, Ziv Wolkowicki, Jason Ziglar, Hong Bae, Thomas Brown,
  Daniel Demitrish, Bakhtiar Litkouhi, Jim Nickolaou, Varsha Sadekar, Wende
  Zhang, Joshua Struble, Michael Taylor, Michael Darms, and Dave Ferguson.
\newblock Autonomous driving in urban environments: Boss and the urban
  challenge.
\newblock \emph{J. Field Robotics}, 25\penalty0 (8):\penalty0 425--466, 2008.

\bibitem[Van Der~Merwe(2004)]{van2004sigma}
Rudolph Van Der~Merwe.
\newblock \emph{Sigma-point Kalman filters for probabilistic inference in
  dynamic state-space models}.
\newblock PhD thesis, University of Stellenbosch, 2004.

\bibitem[\v{S}imandl and Dun{\i}k(2005)]{simandl2005sigma}
M.~\v{S}imandl and J.~Dun{\i}k.
\newblock Sigma point gaussian sum filter design using square root unscented
  filters.
\newblock In \emph{Preprints of the 16th IFAC World Congress}. IFAC. Prague,
  2005.

\bibitem[Wan and Van Der~Merwe(2000)]{wan2000unscented}
Eric~A Wan and Rudolph Van Der~Merwe.
\newblock The unscented kalman filter for nonlinear estimation.
\newblock In \emph{Adaptive Systems for Signal Processing, Communications, and
  Control Symposium 2000. AS-SPCC. The IEEE 2000}, pages 153--158. IEEE, 2000.

\bibitem[Williams and Maybeck(2003)]{williams2003cost}
J.L. Williams and P.S. Maybeck.
\newblock {Cost-function-based gaussian mixture reduction for target tracking}.
\newblock In \emph{Proc. 6th Int. Conf. Inform. Fusion}, volume~2, pages
  1047--1054, 2003.

\bibitem[Zhang et~al.(2003)Zhang, Chen, Sun, and Chan]{GMM_EM_Split_zhang}
Zhihua Zhang, Chibiao Chen, Jian Sun, and Kap~Luk Chan.
\newblock Em algorithms for gaussian mixtures with split-and-merge operation,
  2003.

\bibitem[Ziebart et~al.(2009)Ziebart, Ratliff, Gallagher, Mertz, Peterson,
  Bagnell, Hebert, Dey, and Srinivasa]{ziebart2009planning}
B.D. Ziebart, N.~Ratliff, G.~Gallagher, C.~Mertz, K.~Peterson, J.A. Bagnell,
  M.~Hebert, A.K. Dey, and S.~Srinivasa.
\newblock Planning-based prediction for pedestrians.
\newblock In \emph{Intelligent Robots and Systems, 2009. IROS 2009. IEEE/RSJ
  International Conference on}, pages 3931--3936. IEEE, 2009.

\end{thebibliography}

\end{document}